\documentclass[alpha-refs]{wiley-article}

\usepackage{color}
\usepackage{ulem}

\newcommand{\add}[1]{#1}
\newcommand{\delete}[1]{}
\newcommand{\edit}[2]{#2}
\newcommand{\mnote}[1]{}

\usepackage{siunitx}
\usepackage{csquotes}
\usepackage{cleveref}

\papertype{Research Article}

\title{Development of an offline and online hybrid model for the Integrated Forecasting System}

\author[1\authfn{1}]{Alban Farchi}
\author[2\authfn{1}]{Marcin Chrust}
\author[1\authfn{1}]{Marc Bocquet}
\author[2\authfn{1}]{Massimo Bonavita}

\contrib[\authfn{1}]{Equally contributing authors.}

\affil[1]{CEREA, École des Ponts and EDF R\&D, Île--de--France, France}
\affil[2]{ECMWF, Shinfield Park, Reading, United Kingdom}

\corraddress{Alban Farchin CEREA, École des Ponts and EDF R\&D, Île--de--France, France}
\corremail{alban.farchi@enpc.fr}

\fundinginfo{None.}

\runningauthor{Farchi et al.}

\begin{document}

\maketitle
%\linenumbers
\begin{abstract}
In recent years, there has been significant progress in the development of fully data-driven global numerical weather prediction models. These machine learning weather prediction models have their strength, notably accuracy and low computational requirements, but also their weakness: they struggle to represent fundamental dynamical balances, and they are far from being suitable for data assimilation experiments. Hybrid modelling emerges as a promising approach to address these limitations. Hybrid models integrate a physics-based core component with a statistical component, typically a neural network, to enhance prediction capabilities. In this article, we propose to develop a model error correction for the operational Integrated Forecasting System (IFS) of the European Centre for Medium-Range Weather Forecasts using a neural network. The neural network is initially pre-trained offline using a large dataset of operational analyses and analysis increments. Subsequently, the trained network is integrated into the IFS within the Object-Oriented Prediction System (OOPS) so as to be used in data assimilation and forecast experiments. It is then further trained online using a recently developed variant of weak-constraint 4D-Var. The results show that the pre-trained neural network already provides a reliable model error correction, which translates into reduced forecast errors in many conditions and that the online training further improves the accuracy of the hybrid model in many conditions.

\keywords{data assimilation, machine learning, model error, surrogate model, neural networks, online learning}
\end{abstract}

\section{Introduction}

Since early 2022 and the work of \citet{keisler-2022}, several fully data-driven global numerical weather prediction (NWP) models \citep[notably][]{pathak-2022, bi-2023, lam-2023} have been proposed. These machine learning weather prediction (MLWP) models are based on machine learning methods (neural networks in particular) and are trained using a very large reanalysis dataset of the Earth system, the ERA5 reanalysis \citep{hersbach-2020} produced by the European Centre for Medium-Range Weather Forecasts (ECMWF). The advantage of using reanalyses as training dataset of the MLWP models is that they are in general more accurate than raw observations. More importantly, it circumvents the issue that Earth observations are always sparse (i.e., not available on a regular computational grid) and often indirectly related to the forecast quantities of interest, which is problematic for standard machine learning approaches. Despite being a technological breakthrough --- \add{deterministic }MLWP models are able to produce forecasts with an accuracy arguably competitive with the best physics-based NWP models at a fraction of the computational time and cost --- they also have \delete{their }weaknesses. First, data-driven models are trained using a reanalysis dataset, which, even though it represents our best knowledge of the state of the Earth system over the past, is still affected by biases which mainly come from the use of a physics-based model in the assimilation system used to produce the reanalysis. A solution to mitigate the biases would be to update the reanalysis dataset using the trained MLWP model, effectively combining data assimilation and machine learning as originally proposed by \citet{brajard-2020, bocquet-2020}. However, the existing MLWP models are by construction designed for forecasting tasks and are far from being suitable for assimilation purposes \citep{bocquet-2023, bonavita-2024}\add{, because their horizontal and vertical spatial and temporal resolutions are not sufficient, but more importantly because they lack physical consistency}. Beyond these considerations, they also present some limitations in the forecast applications. In particular, they tend to produce progressively smoother predictions, which can be seen as a consequence of the \enquote{double penalty effect} (these models are usually trained using mean squared or mean absolute error as loss function). Consequently, they struggle to represent fundamental dynamical balances in the atmosphere such as geostrophic/ageostrophic flows and divergent/rotational winds as illustrated by \citet{bonavita-2024}. \add{Generative MLWP models have recently been proposed as a potential way to circumvent these issues \citep{price-2024, finn-2024a} by relying on ensembles.} Hybrid modelling, in other words using a physics-based core model supplemented by a data-driven component, can be seen as \edit{a}{another} potential solution to overcome, or at least mitigate, the limitations of both traditional NWP models and \edit{new}{deterministic} MLWP models.

Hybrid modelling is by construction closely related to model error correction as the purpose of the data-driven (or statistical) component is precisely to correct the errors of the physics-based core component. Hybrid modelling or model error correction is an active area of research with contributions from both the data assimilation and the machine learning community. From a data assimilation perspective, an exemplar is the development of weak-constraint methods, that is, data assimilation methods relaxing the perfect model assumption \citep{tremolet-2006}, and in particular the iterative ensemble Kalman filter in the presence of additive noise \citep{sakov-2018} in statistical data assimilation, and of the forcing formulation of weak-constraint 4D-Var \citep{laloyaux-2020} in variational data assimilation. In the machine learning community, researchers are more and more inclined to acknowledge the value of physics-based models, which are based on decades of experience and knowledge in numerical modelling \citep{levine-2022}. Even though hybrid models can be more difficult to implement than surrogate models, they are often more accurate while reducing data demands \citep{watson-2019,farchi-2021}. There are many examples of hybrid modelling in the geosciences, ranging from data-driven subgrid scale parametrisations \citep{rasp-2018, bolton-2019, gagne-2020, finn-2023, ross-2023} to generic model error correction \citep{bonavita-2020, wikner-2020, farchi-2021, farchi-2021b, brajard-2020b, chen-2022} and super resolution \citep{barthelemy-2022}.

For practical reasons, hybrid models are usually trained offline\add{, i.e. once the entire observation dataset is available}. Online approaches\add{, i.e. improving the models as new observations become available,} have however attractive advantages over offline approaches. \add{Online approaches are of course not limited to hybrid modelling and can in principle be used for a wide range of cases, from sub-grid scale parametrisation to full model emulation, as illustrated for example by \citet{bocquet-2020a}.} \edit{They}{In any case, they} have better synergies with data assimilation methods and online trained models are usually more accurate \citep{farchi-2021b, farchi-2023}. For these reasons, there is a growing interest for online approaches in hybrid modelling, even though they are significantly more difficult to implement. Above all, online training usually requires the adjoint operator of the physics-based model to correct. This is not a problem within an auto-differentiable framework \citep[e.g., ][]{farchi-2021b, frezat-2022, levine-2022, kochkov-2023} but NWP \edit{models}{codes} are rarely auto-differentiable. Yet, recognising that online training is very similar to parameter estimation in data assimilation, several examples of online learning methods have recently emerged in both statistical \citep{bocquet-2020, rasp-2020a, gottwald-2021, lopez-gomez-2022} and variational data assimilation \citep{farchi-2021b}. The latter method, called neural network formulation of weak-constraint 4D-Var (NN 4D-Var), has been simplified by \citet{farchi-2023} and implemented in the Object-Oriented Prediction System (OOPS) developed at ECMWF.% \add{[line break added here]}

In the present article, our objective is to push forward this effort and demonstrate that, after successful applications to low-order and intermediate models, NN 4D-Var \edit{is ready for}{can be used to build hybrid models on top of} realistic, state-of-the-art prediction systems like the ECMWF Integrated Forecasting System \citep[IFS,][]{bonavita-2017} within OOPS. Building on the preliminary work of \citet{bonavita-2020}, we pre-train offline a neural network to correct model error in the IFS using a large dataset of analyses and analysis increments. The network is then embedded in the IFS so as to be used in data assimilation experiments, in particular with NN 4D-Var within online training experiments. The article is structured as follows. \Cref{sec:methodology} presents the methodological aspects of NN 4D-Var and the two-step (offline then online) training process. The offline training step is described and illustrated in \cref{sec:offline}, while \cref{sec:online} focuses on the online training step and its results. Finally, the results are discussed in \cref{sec:discussion} and conclusions and perspectives are given in \cref{sec:conclusions}.

\section{Methodology}
\label{sec:methodology}

\add{In this section, we introduce the main methodological aspects of the present work, which are the same as in our previous work \citep{farchi-2023}.}

\subsection{Strong-constraint 4D-Var}

Let us consider a standard, discrete time data assimilation problem, whose goal is to follow the evolution of the system using sparse and noisy observations. With variational techniques, for example 4D-Var \citep{courtier-1994}, the observations $\left(\mathbf{y}_{0}, \ldots, \mathbf{y}_{L}\right)$ between times $t_0$ and $t_k$ are assimilated by minimising the cost function
\begin{equation}
    \label{eq:methodology-sc4dvar-cost-gaussian}
    \mathcal{J}^{\mathsf{sc}}\left(\mathbf{x}_{0}\right) \triangleq \frac{1}{2} \big\|\mathbf{x}_{0}-\mathbf{x}^{\mathsf{b}}_{0}\big\|^{2}_{\mathbf{B}^{-1}} + \frac{1}{2} \sum_{k=0}^{L} \big\|\mathbf{y}_{k}-\boldsymbol{\mathcal{H}}_{k}\circ\boldsymbol{\mathcal{M}}_{k:0}\left(\mathbf{x}_{0}\right)\big\|^{2}_{\mathbf{R}^{-1}_{k}},
\end{equation}
where the notation $\left\|\mathbf{v}\right\|^{2}_{\mathbf{M}}$ stands for the squared Mahalanobis norm $\mathbf{v}^{\top}\mathbf{M}\mathbf{v}$, and where the window length is $L$. This cost function corresponds to the negative log-likelihood $-\ln p\left(\mathbf{x}_{0}|\mathbf{y}_{0}, \ldots, \mathbf{y}_{L}\right)$ in the Gaussian case where the background error follows a centred normal distribution with covariance matrix $\mathbf{B}$ and the observation errors follow centred normal distributions with covariance matrices $\mathbf{R}_{k}$.

In this equation, $\boldsymbol{\mathcal{M}}_{k:l}:\mathbb{R}^{N_{\mathsf{x}}}\to\mathbb{R}^{N_{\mathsf{x}}}$ is the resolvent of the dynamical model from $t_l$ to $t_k$, which is used to propagate the system state in time:
\begin{equation}
    \label{eq:methodology-sc4dvar-model}
    \mathbf{x}_{k} = \boldsymbol{\mathcal{M}}_{k:l}\left(\mathbf{x}_{l}\right),
\end{equation}
and $\boldsymbol{\mathcal{H}}_{k}:\mathbb{R}^{N_{\mathsf{x}}}\to\mathbb{R}^{N_{\mathsf{y}}}$ is the observation operator at $t_{k}$, which is used to represent the observation process:
\begin{equation}
    \label{eq:methodology-sc4dvar-observations}
    \mathbf{y}_{k} = \boldsymbol{\mathcal{H}}_{k}\left(\mathbf{x}_{k}\right) + \mathbf{v}_{k}, \quad \mathbf{v}_{k}\sim\mathcal{N}\left(\mathbf{0}, \mathbf{R}\right).
\end{equation}

The analysis at the start of the window $\mathbf{x}^{\mathsf{a}}_{0}$ is obtained by minimising the cost function $\mathcal{J}^{\mathsf{sc}}$ and is then propagated until the start of the next window to provide the next background state $\mathbf{x}^{\mathsf{b}}_{0}$. This approach is called \emph{strong-constraint} 4D-Var because it assumes that the model \cref{eq:methodology-sc4dvar-model} is perfect.

\subsection{Weak-constraint 4D-Var: a neural network variant}
\label{ssec:methodology-nn4dvar}

Model error is one of the main limitations of all current data assimilation algorithms. In the past few years, several approaches have been developed to correct model errors or at least to mitigate their impact in data assimilation experiments \citep{sakov-2018, laloyaux-2020, laloyaux-2020b}. In the present work, we are going to use the approach initially derived by \citet{farchi-2021b} and then adapted to the incremental 4D-Var formulation by \citet{farchi-2023}.

Within this approach, the perfect model evolution \cref{eq:methodology-sc4dvar-model} is replaced with
\begin{equation}
    \label{eq:methodology-nn4dvar-model}
    \mathbf{x}_{k} = \boldsymbol{\mathcal{M}}_{k:k-1}\left(\mathbf{x}_{k-1}\right) + \mathbf{w}, \quad \mathbf{w} = \boldsymbol{\mathcal{F}}\left(\mathbf{p}, \mathbf{x}_{0}\right),
\end{equation}
where $\boldsymbol{\mathcal{F}}$ is a statistical model, typically a neural network, parametrised by $\mathbf{p}$. Let us denote $\boldsymbol{\mathcal{M}}^{\mathsf{h}}_{k:0}\left(\mathbf{p}, \mathbf{x}_{0}\right)$ the time integration between $t_0$ and $t_k$, where the superscript $\mathsf{h}$ is used to emphasise the fact that the model is now hybrid, with a physical part ($\boldsymbol{\mathcal{M}}$) supplemented by a statistical part ($\boldsymbol{\mathcal{F}}$). Effectively, the neural network $\boldsymbol{\mathcal{F}}$ can be seen as a model of the model error of the physical model $\boldsymbol{\mathcal{M}}$.

The original strong-constraint 4D-Var is then modified in two different ways. First, the physical model $\boldsymbol{\mathcal{M}}$ is replaced by the hybrid model $\boldsymbol{\mathcal{M}}^{\mathsf{h}}$. Second, the parameters of the statistical model $\mathbf{p}$ are included in the control variable, so that they can be estimated as part of the data assimilation analysis. The resulting cost function reads:
\begin{equation}
    \label{eq:methodology-nn4dvar-cost-gaussian}
    \mathcal{J}^{\mathsf{nn}}\left(\mathbf{p}, \mathbf{x}_{0}\right) \triangleq \frac{1}{2} \big\|\mathbf{x}_{0}-\mathbf{x}^{\mathsf{b}}_{0}\big\|^{2}_{\mathbf{B}^{-1}} + \frac{1}{2} \big\|\mathbf{p}-\mathbf{p}^{\mathsf{b}}\big\|^{2}_{\mathbf{P}^{-1}} + \frac{1}{2} \sum_{k=0}^{L} \big\|\mathbf{y}_{k}-\boldsymbol{\mathcal{H}}_{k}\circ\boldsymbol{\mathcal{M}}^{\mathsf{h}}_{k:0}\left(\mathbf{p}, \mathbf{x}_{0}\right)\big\|^{2}_{\mathbf{R}^{-1}_{k}},
\end{equation}
where we have assumed that the background errors for model state and model parameters are independent and follow centred normal distributions with covariance matrices $\mathbf{B}$ and $\mathbf{P}$, respectively. 

As for strong-constraint 4D-Var, the cost function $\mathcal{J}^{\mathsf{nn}}$ is minimised to obtain the analysis at the start of the window, for both model state ($\mathbf{x}^{\mathsf{a}}_{0}$) and model parameters ($\mathbf{p}^{\mathsf{a}}$). The analysis is then propagated until the start of the next window, with the hybrid model, to get a value for the next background state $\mathbf{x}^{\mathsf{b}}_{0}$. For the model parameters, we assume that there is no evolution, i.e. the next background parameters $\mathbf{p}^{\mathsf{b}}$ are equal to the current $\mathbf{p}^{\mathsf{a}}$. This approach is a parametrised variant of weak-constraint 4D-Var, and has been called neural network formulation of weak-constraint 4D-Var (NN 4D-Var) in the case where $\boldsymbol{\mathcal{F}}$ is a neural network and $\mathbf{p}$ contains the weights and biases of this neural network.

In the original formulation of \citet{farchi-2021b}, the model bias $\mathbf{w}$ is recomputed at each model \edit{update}{step from $t_{k}$ to $t_{k+1}$} using $\boldsymbol{\mathcal{F}}$. Here, we use the simplified formulation, where $\mathbf{w}$ is computed only once, at the start of the window. \edit{As shown by Farchi et al. (2023), this}{This} simplification can be utilised to build the new 4D-Var variant on top of the existing forcing formulation of weak-constraint 4D-Var adopted in the IFS \citep{laloyaux-2020}. \add{More technical details can be found in \citet{farchi-2023}, in particular the pseudo-code that is used to compute the gradient of the incremental cost function \citep[Algorithm 3 of ][]{farchi-2023}.}

\subsection{A two-step training process: offline then online}
\label{ssec:two-step-training}

The quality of the NN 4D-Var analysis critically depends on the choice of the background values of the model parameters $\mathbf{p}^{\mathsf{b}}$ and background error covariance matrix for model parameters $\mathbf{P}$. Let us see how $\mathbf{p}^{\mathsf{b}}$ and $\mathbf{P}$ can be chosen.

Without further knowledge on the structure of the model parameters, a simple choice for the background error covariance matrix is $\mathbf{P}=p^{2}\mathbf{I}$, where $p$ is the standard deviation and $\mathbf{I}$ is the identity matrix. The merits of this choice are discussed in details by \citet{farchi-2021b}.

In a cycled data assimilation context, the background for model parameters $\mathbf{p}^{\mathsf{b}}$ is given by the model parameter analysis $\mathbf{p}^{\mathsf{a}}$ of the previous data assimilation cycle assuming persistence. Therefore, the background is progressively updated over the cycles, and we just need to provide the background for the very first cycle. In principle, we can provide any initial background, as long as the standard deviation $p$ of the background error covariance matrix is sufficiently large. For example, \citet{farchi-2021b} have used random values for the initial background $\mathbf{p}^{\mathsf{b}}$, with a standard deviation $p$ larger in the first cycles to account for the fact that at the start of the experiment, the model parameters are poorly known. Alternatively, the neural network $\mathcal{F}$ can be pretrained offline using a large dataset of analyses and analysis increments \citep{farchi-2021}, thus providing a value for $\mathbf{p}^{\mathsf{b}}$. The advantage of this approach, advocated for example in \citet{farchi-2023}, is that we avoid a cold start of the neural network training, which could lead to immediate divergence. In that case, the neural network training can be seen as a two-step process, where the network is first trained offline using a large dataset and then online within the data assimilation cycles. This is the approach that we will follow here.

\section{Step 1: offline training}
\label{sec:offline}

\subsection{Description of the training dataset}
\label{ssec:offline-dataset-description}

The objective of this work is to provide a model error correction for the Integrated Forecasting System (IFS) developed at the European Centre for Medium-range Weather Forecasts (ECMWF). Therefore, for the offline training step, we will use a dataset gathering all the operational analyses and background forecasts produced by ECMWF between 01/01/2017 and 01/10/2021. Even though it covers multiple IFS cycles (from 43r1 to 47r2), there was no major model update in this period so that the model error is expected be roughly similar throughout the entire dataset. Note that most of the dataset has been produced using strong constraint 4D-Var, with the exception of the last 16 months, produced using weak-constraint 4D-Var (which was introduced in June 2020 with cycle 47r1). We initially thought that mixing strong constraint and weak constraint 4D-Var would have a limited impact on the results. This is further discussed in \cref{sapp:offline-spatial-scores}.

In the end, there is a total of $\num{1734}$ days in the dataset, which are partitioned as follows. The first \edit{$\num{1370}$}{$N^{\mathsf{train}}_{\mathsf{day}}=\num{1370}$} days, from 01/01/2017 to 01/10/2020, form the training set. The last $\num{364}$ days are distributed between validation and testing set by batch: the first $\num{4}$ days are discarded, the following $\num{8}$ days are put in the validation set, the following $\num{4}$ days are discarded, the following $\num{8}$ days are put in the testing set, and this process is repeated until the end of the data. This gives us a total of $\num{121}$ days in the validation set and $\num{120}$ days in the testing set, arranged into 15 batches of consecutive days. We choose this method (i) to ensure that the validation and testing data are posterior to the training data, and (ii) to have a representation (at least partial) of a full year, and hence of seasonality effects, in both the validation and testing data without having to set aside two entire years. For completeness, throughout the entire dataset the data assimilation window length is $\num{12}$ hours, in such a way that for each day of dataset, we have exactly two state snapshots. In the following, the training, validation, and testing sets will be called $\mathcal{T}_{\mathsf{train}}$, $\mathcal{T}_{\mathsf{valid}}$, and $\mathcal{T}_{\mathsf{test}}$. \add{Note that the sensitivity to the size of the training dataset is illustrated in \cref{app:offline-size-sensitivity}.}

In this dataset, four variables are selected: logarithm of surface pressure (lnsp), temperature (t), vorticity (vo), and divergence (d). For the latter three variables, we keep all $\num{137}$ levels, which means that at each latitude-longitude grid point, we have a total of \edit{$\num{1}+\num{3}\times\num{137}=\num{412}$}{$N_{\mathsf{var}}=\num{1}+\num{3}\times\num{137}=\num{412}$} variables. Furthermore, the original data is archived in spectral space. For the present work, we retrieved the data at the intermediate T63 resolution, where T means that we have used a triangular spectral truncation. In order to be able to perform extensive tests, most of the offline experiments are performed at the coarse T15 resolution. An example of training at higher resolution (namely T31 resolution) is illustrated in \cref{app:offline-hr}. Note that this choice is consistent with the conclusion of previous papers that only large-scale model errors are predictable \citep{laloyaux-2020, laloyaux-2020b, bonavita-2020}. In the T15 resolution, for each of the \edit{$\num{412}$}{$N_{\mathsf{var}}=\num{412}$} state variables, there are $\num{16}\times\num{17}/\num{2}=\num{136}$ complex degrees of freedom\footnote{In other words, the number of independent complex spectral coefficients at T15 resolution is $\num{136}$.} or equivalently \edit{$\num{272}$}{$N_{\mathsf{spec}}=\num{272}$} real degrees of freedom, for a total of \edit{$\num{272}\times\num{412}=\num{112064}$}{$N_{\mathsf{spec}}\times N_{\mathsf{var}}=\num{112064}$} real degrees of freedom per state snapshot.

Finally, in these offline training experiments, we would like to target the analysis increments (analysis minus background at the start of each window) as they can be seen as a proxy for model error \citep{farchi-2021}. For this task, two strategies have emerged:
\begin{itemize}
    \item In the first approach, the state predictor is the analysis at the start of the previous window \citep[e.g.,][]{farchi-2021, brajard-2020b}. In other words, the neural network should emulate the map
    \begin{equation}
        \label{eq:offline-pred}
        \mathbf{x}^{\mathsf{a}}_{0}\left(t\right) \mapsto \mathbf{x}^{\mathsf{a}}_{0}\left(t+1\right) - \mathbf{x}^{\mathsf{b}}_{0}\left(t+1\right),
    \end{equation}
    where the $0$ subscript indicates that the quantities are extracted at the start of the window and $\left(t\right)$ and $\left(t+1\right)$ refer to the $t$-th and $(t+1)$-th window, respectively. In the present work, this approach will be called \emph{prediction} mode to emphasise the time lag between input and output.
    \item In the second approach, the state predictor is the background of the current window \citep[e.g.,][]{bonavita-2020, laloyaux-2022}. This time, the neural network should emulate the map
    \begin{equation}
        \label{eq:offline-pproc} \mathbf{x}^{\mathsf{b}}_{0}\left(t\right) \mapsto \mathbf{x}^{\mathsf{a}}_{0}\left(t\right) - \mathbf{x}^{\mathsf{b}}_{0}\left(t\right).
    \end{equation}
    In the present work, this approach will be called \emph{post-processing} mode.
\end{itemize}
Of course, it is also possible to combine the two approaches \citep{finn-2023}, but in order to stay within the framework presented in \cref{ssec:methodology-nn4dvar}, we need at most one state predictor, which is why we restrict our study to the prediction and post-processing modes. On one hand, the time lag between input and output means that the inference problem should be more complex in prediction mode than in post-processing mode. On the other hand, we believe that training a neural network in prediction mode should result in a more accurate hybrid model as formulated in \cref{ssec:methodology-nn4dvar}. One of the objective of the present work is to validate this hypothesis.

\subsection{Neural network architecture}
\label{ssec:offline-nn-architecture}

For practical reasons, even though the data are available in spectral space, we choose to apply the neural network in grid-point space. Therefore, the entire dataset described in \cref{ssec:offline-dataset-description} has to be interpolated onto a grid. For the present study, we choose to use a rectangular Gauss--Legendre grid with $N_{\mathsf{lat}}=\num{16}$ latitude nodes (distributed according to the zeros of the Legendre Polynomial of degree $\num{16}$) and $N_{\mathsf{lon}}=\num{31}$ longitude nodes (equally distributed). This grid is the smallest possible grid that can represent a field in the T15 resolution. Note, however, that there are, for one field, \edit{$\num{496}$}{$N_{\mathsf{lat}}\times N_{\mathsf{lon}}=\num{496}$} grid nodes, about double the number of degrees of freedom (\edit{$\num{272}$}{$N_{\mathsf{spec}}=\num{272}$}). Therefore, there is redundant information in the interpolated data, which is unavoidable with rectangular grids. Nevertheless, rectangular grids have the advantage of being easier to manipulate, which is why we choose to keep a rectangular grid and to compensate oversampling in the polar regions by using Gauss--Legendre weights, as will be explained later.

In this context, we choose to use a vertical/column architecture, where the same neural network is applied independently to each atmospheric column. At first sight, this approach can be seen quite restrictive because it ignores horizontal spatial relationships, in particular those between neighbouring grid points. Nevertheless, it is based on the intuition that in global weather forecast models, the majority of the errors comes from the parameterisations of the physical processes (which are generally implemented in columns and typically only account for vertical processes) and not from the dynamical core. Furthermore, this approach has many practical advantages, which is why it has already been applied to model error correction of large-scale weather forecast models \citep[e.g.,][]{bonavita-2020, chen-2022, kochkov-2023} with reasonable success, in particular:
\begin{itemize}
    \item We reduce the dimension of the input and output space of the neural network from \edit{$\num{412}\times\num{16}\times\num{31}=\num{204352}$}{$N_{\mathsf{var}}\times N_{\mathsf{lat}}\times N_{\mathsf{lon}}=\num{204352}$} to \edit{$\num{412}$}{$N_{\mathsf{var}}=\num{412}$}, in such a way that the choice of the neural network and its training step will be relatively easy (from a technical point of view) and quick.
    \item The trained neural network will be independent of the choice of the grid. In particular, it can be trained in the \edit{$\num{16}\times\num{31}$}{$N_{\mathsf{lat}}\times N_{\mathsf{lon}}=\num{16}\times\num{31}$} Gauss--Legendre grid and later used in any other grid.
    \item Later on, in the online experiments, different atmospheric columns may be stored in memory of distinct processors. With a column architecture, the neural network can be applied without the need for message passing interface between processors.
\end{itemize}
In the present work, we choose to use an \enquote{all-in, all-out} approach, where we train a single neural network for all four variables (lnsp, t, vo, d) in the atmospheric column, for two reasons. First, this enables the neural network to utilise cross-correlation between variables, resulting in a potentially more accurate correction. Second, this will make the online implementation easier.

In order to be able to capture additional spatio-temporal patterns (for example, the effect of seasonality) we add to the input space a total of $\num{8}$ extra predictors, namely the sinus and \edit{cosine}{cosinus} of (i) latitude, (ii) longitude, (iii) time of the day, and (iv) day of the year\footnote{\add{Note that we do not make a distinction between normal years and leap years. In practice the day of the year index ranges from $0$ to $364$ for normal years and from $0$ to $365$ for leap years.}}. Consequently, the dimension of the input and output space of the neural network are finally set to \edit{$\num{420}$}{$N_{\mathsf{in}}=N_{\mathsf{var}}+\num{8}=\num{420}$} and \edit{$\num{412}$}{$N_{\mathsf{out}}=N_{\mathsf{var}}=\num{412}$}, respectively.

\begin{figure}[tbp]
    \centering
    \includegraphics[width=\linewidth]{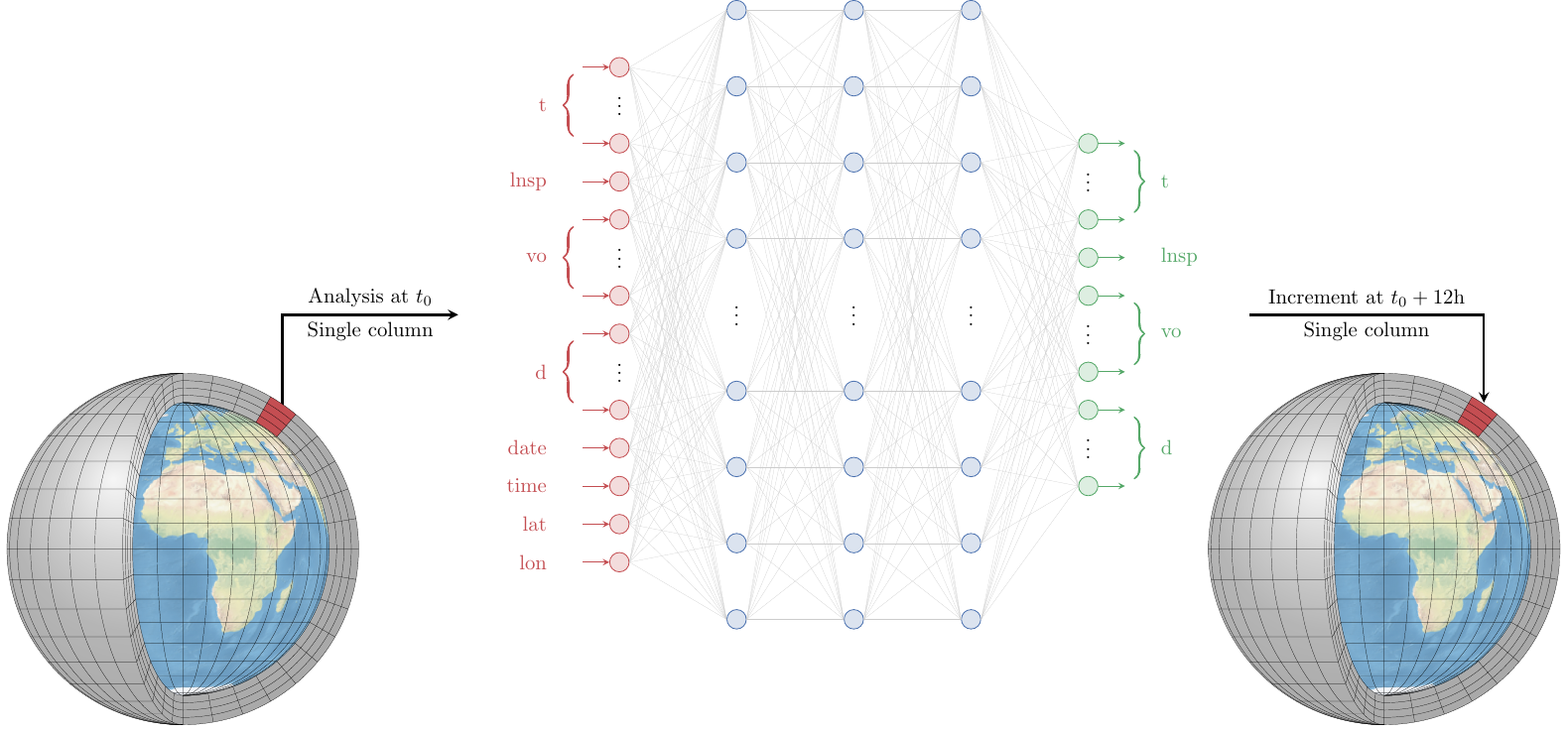}
    \caption{Illustration of the column neural network architecture used in the present work.}
    \label{fig:offline-nn}
\end{figure}

After preliminary screening experiments (not described here), we decided to use a feed-forward fully-connected neural network, made up of four internal (hidden) layers with $\num{512}$ neurons each and with the tanh activation function and one output linear layer with $\num{412}$ neurons (one for each output variable), for a total of $\num{1214876}$ parameters. For comparison, in the training dataset there is a total of \edit{$\num{1370}\times\num{2}\times\num{412}\times{16}\times{31} = \num{559924480}$}{$N^{\mathsf{train}}_{\mathsf{day}}\times\num{2}\times N_{\mathsf{out}}\times N_{\mathsf{lat}}\times N_{\mathsf{lon}} = \num{559924480}$} \edit{data points}{floating point numbers} (number of training dates times number of windows per date times dimension of the output times number of latitude nodes times number of longitude nodes), i.e. two orders of magnitude more than the number of parameters. \delete{Therefore, we do not expect to encounter significant overfitting issues. }This neural network is depicted in \cref{fig:offline-nn}. Alternative architectures are discussed in \cref{ssec:conclusions-architecture}.

Finally, let us mention that the data is normalised before being sent to the neural network. In practice, both the input and the output of the neural network are standardised (subtracting the mean and dividing by the standard deviation), using independent normalisation coefficients for each of the \edit{$\num{412}$}{$N_{\mathsf{var}}=\num{412}$} input and \edit{$\num{412}$}{$N_{\mathsf{var}}=\num{412}$} output variables -- the $\num{8}$ extra predictors are not normalised -- computed using the training set only.

\subsection{Loss function}

Let us start by defining some notation:
\begin{itemize}
    \item Let \edit{$\mathbf{x}^{\mathsf{i}}_{t}\in\mathbb{R}^{420\times16\times31}$}{$\mathbf{x}^{\mathsf{i}}_{t}\in\mathbb{R}^{N_{\mathsf{in}}\times N_{\mathsf{lat}}\times N_{\mathsf{lon}}}$} and \edit{$\mathbf{x}^{\mathsf{o}}_{t}\in\mathbb{R}^{412\times16\times31}$}{$\mathbf{x}^{\mathsf{o}}_{t}\in\mathbb{R}^{N_{\mathsf{out}}\times N_{\mathsf{lat}}\times N_{\mathsf{lon}}}$} be the $t$-th input and output states in the dataset in grid-point space. Following \cref{eq:offline-pred,eq:offline-pproc}, $\mathbf{x}^{\mathsf{i}}_{t}=\mathbf{x}^{\mathsf{a}}_{0}\left(t\right)$ and $\mathbf{x}^{\mathsf{o}}_{t}=\mathbf{x}^{\mathsf{a}}_{0}\left(t+1\right)-\mathbf{x}^{\mathsf{b}}_{0}\left(t+1\right)$ in prediction mode, whereas $\mathbf{x}^{\mathsf{i}}_{t}=\mathbf{x}^{\mathsf{b}}_{0}\left(t\right)$ and $\mathbf{x}^{\mathsf{o}}_{t}=\mathbf{x}^{\mathsf{a}}_{0}\left(t\right)-\mathbf{x}^{\mathsf{b}}_{0}\left(t\right)$ in post-processing mode.
    \item Let \edit{$\mathbf{z}^{\mathsf{i}}_{t}\in\mathbb{R}^{420\times16\times31}$}{$\mathbf{z}^{\mathsf{i}}_{t}\in\mathbb{R}^{N_{\mathsf{in}}\times N_{\mathsf{lat}}\times N_{\mathsf{lon}}}$} and \edit{$\mathbf{z}^{\mathsf{o}}_{t}\in\mathbb{R}^{412\times16\times31}$}{$\mathbf{z}^{\mathsf{o}}_{t}\in\mathbb{R}^{N_{\mathsf{out}}\times N_{\mathsf{lat}}\times N_{\mathsf{lon}}}$} be the normalised counterparts of $\mathbf{x}^{\mathsf{i}}_{t}$ and $\mathbf{x}^{\mathsf{o}}_{t}$, using the normalisation described at the end of \cref{ssec:offline-nn-architecture}.
    \item Let \edit{$\mathbf{z}^{\mathsf{i}}_{t, i, j}\in\mathbb{R}^{420}$}{$\mathbf{z}^{\mathsf{i}}_{t, i, j}\in\mathbb{R}^{N_{\mathsf{in}}}$} and \edit{$\mathbf{z}^{\mathsf{o}}_{t, i, j}\in\mathbb{R}^{412}$}{$\mathbf{z}^{\mathsf{o}}_{t, i, j}\in\mathbb{R}^{N_{\mathsf{out}}}$} be the vertical profiles of $\mathbf{z}^{\mathsf{i}}_{t}$ and $\mathbf{z}^{\mathsf{o}}_{t}$ at the node defined by the $i$-th latitude and the $j$-th longitude.
    \item Let $\widehat{\boldsymbol{\mathcal{G}}}$ be the column neural network in the normalised grid-point space (i.e. acting on $\mathbf{z}^{\mathsf{i}}_{t, i, j}$).
    \item Let $\boldsymbol{\mathcal{G}}$ be the column neural network in the non-normalised grid-point space (i.e. acting on $\mathbf{x}^{\mathsf{i}}_{t, i, j}$), which corresponds to composing $\widehat{\boldsymbol{\mathcal{G}}}$ with the appropriate normalisation and denormalisation operators.
    \item Let $\widehat{\boldsymbol{\mathcal{F}}}$ be the full neural network in the normalised grid-point space (i.e. acting on $\mathbf{z}^{\mathsf{i}}_{t}$), which corresponds to $\widehat{\boldsymbol{\mathcal{G}}}$ operating over all \edit{$16\times 31$}{$N_{\mathsf{lat}}\times N_{\mathsf{lon}}$} grid nodes.
    \item Let $\boldsymbol{\mathcal{F}}$ be the full neural network in the non-normalised grid-point space (i.e. acting on $\mathbf{x}^{\mathsf{i}}_{t}$), which corresponds to composing $\widehat{\boldsymbol{\mathcal{F}}}$ with the appropriate normalisation and denormalisation operators, or equivalently to $\boldsymbol{\mathcal{G}}$ operating over all \edit{$16\times 31$}{$N_{\mathsf{lat}}\times N_{\mathsf{lon}}$} grid nodes.
\end{itemize}
With these notations, for a set of well-calibrated parameters $\mathbf{p}^{\star}$, we would want:
\begin{subequations}
\begin{align}
    \mathbf{z}^{\mathsf{o}}_{t, i, j} &\approx \widehat{\boldsymbol{\mathcal{G}}}\left(\mathbf{p}^{\star}, \mathbf{z}^{\mathsf{i}}_{t, i, j}\right),&
    \mathbf{x}^{\mathsf{o}}_{t, i, j} &\approx\boldsymbol{\mathcal{G}}\left(\mathbf{p}^{\star}, \mathbf{x}^{\mathsf{i}}_{t, i, j}\right),\\
    \mathbf{z}^{\mathsf{o}}_{t} &\approx \widehat{\boldsymbol{\mathcal{F}}}\left(\mathbf{p}^{\star}, \mathbf{z}^{\mathsf{i}}_{t}\right).&
    \mathbf{x}^{\mathsf{o}}_{t} &\approx \boldsymbol{\mathcal{F}}\left(\mathbf{p}^{\star}, \mathbf{x}^{\mathsf{i}}_{t}\right).
\end{align}
\end{subequations}

In the offline training step, the parameters $\mathbf{p}$ are found by minimising the following weighted mean-squared error (wMSE):
\begin{equation}
    \widehat{\mathcal{L}}\left(\mathbf{p}\right) \triangleq \sum_{t\in\mathcal{T}_{\mathsf{train}}} \sum_{i=1}^{N_{\mathsf{lat}}}\sum_{j=1}^{N_{\mathsf{lon}}} w_{i}\left\| \mathbf{z}^{\mathsf{o}}_{t, i, j} - \widehat{\boldsymbol{\mathcal{G}}}\left(\mathbf{p}, \mathbf{z}^{\mathsf{i}}_{t, i, j}\right)\right\|^{2},
\end{equation}
where $w_{i}$ is the Gauss--Legendre weight at latitude $i$, $\left\|.\right\|$ is the standard L$^{2}$-norm. For simplicity the normalisation constant has been dropped. By construction of the Gauss--Legendre weights, this loss, computed in grid-point space, should be very close to the equivalent loss in spectral space
\begin{equation}
    \widehat{\mathcal{L}}_{\mathsf{spec}}\left(\mathbf{p}\right) \triangleq \sum_{t\in\mathcal{T}_{\mathsf{train}}} \left\| \mathbf{S}\mathbf{z}^{\mathsf{o}}_{t} - \mathbf{S}\widehat{\boldsymbol{\mathcal{F}}}\left(\mathbf{p}, \mathbf{z}^{\mathsf{i}}_{t}\right)\right\|^{2},
\end{equation}
where \edit{$\mathbf{S}:\mathbb{R}^{412\times16\times31}\to\mathbb{R}^{412\times272}$}{$\mathbf{S}:\mathbb{R}^{N_{\mathsf{out}}\times N_{\mathsf{lat}}\times N_{\mathsf{lon}}}\to\mathbb{R}^{N_{\mathsf{out}}\times272}$} is the transformation from grid-point to spectral space (applied independently for each variable and each vertical level). The residual difference between $\widehat{\mathcal{L}}\left(\mathbf{p}\right)$ and $\widehat{\mathcal{L}}_{\mathsf{spec}}\left(\mathbf{p}\right)$ comes from the additional degrees of freedom in grid-point space, which cannot be represented at the T15 resolution.

\subsection{Neural network training and results}
\label{ssec:offline_training}

In both prediction and post-processing mode, the neural network is trained for a maximum of $\num{2048}$ epochs using Adam algorithm \citep{kingma-2015} with a batch size of $\num{2048}$ and a relatively small learning rate of $5\times 10^{-5}$. In addition, for each hidden layer, the dropout technique is used with a rate of $\num{0.1}$. The batch size may seem really large, but one has to keep in mind that it is counted in number of vertical profiles. Indeed, with our \edit{$\num{16}\times\num{31}$}{$N_{\mathsf{lat}}\times N_{\mathsf{lon}}=\num{16}\times\num{31}$} grid, $\num{2048}$ profiles correspond to approximately $\num{4}$ entire state snapshots. Furthermore, we use an early stopping callback on the validation loss with a patience of $\num{128}$ epochs. After triggering the early stopping callback, we restore the optimal parameters.

Now that the network has been trained by minimising $\widehat{\mathcal{L}}\left(\mathbf{p}\right)$, we evaluate it using the following relative wMSE:
\begin{equation}
    \label{eq:offline-normalised-score}\mathcal{S}_{\mathsf{var}}\left(\mathbf{p}\right) \triangleq \frac{\sum\limits_{t\in\mathcal{T}_{\mathsf{test}}} \sum\limits_{i=1}^{N_{\mathsf{lat}}}\sum\limits_{j=1}^{N_{\mathsf{lon}}} w_{i}\left\| \mathbf{x}^{\mathsf{o}}_{t, i, j} - \boldsymbol{\mathcal{G}}\left(\mathbf{p}, \mathbf{x}^{\mathsf{i}}_{t, i, j}\right)\right\|^{2}_{\mathsf{var}}}{\sum\limits_{t\in\mathcal{T}_{\mathsf{test}}} \sum\limits_{i=1}^{N_{\mathsf{lat}}}\sum\limits_{j=1}^{N_{\mathsf{lon}}} w_{i}\left\| \mathbf{x}^{\mathsf{o}}_{t, i, j}\right\|^{2}_{\mathsf{var}}},
\end{equation}
where $\left\|.\right\|^{2}_{\mathsf{var}}$ corresponds to the standard L$^{2}$-norm, but computed only on the subspace corresponding to variable $\mathsf{var}$ (which can be lnsp, t, vo, or d). The advantage of this score is that (i) it is independent from the normalisation since it is computed in the non-normalised grid-point space, and (ii) it is easily interpretable:
\begin{itemize}
    \item $\mathcal{S}_{\mathsf{var}}\left(\mathbf{p}\right)=1$ when the predictions are always $\mathbf{0}$ (no correction);
    \item $\mathcal{S}_{\mathsf{var}}\left(\mathbf{p}\right)\leq1$ if the predictions are on average better than having no correction;
    \item $\mathcal{S}_{\mathsf{var}}\left(\mathbf{p}\right)=0$ when the predictions are perfect.
\end{itemize} 

\begin{table}[tbp]
    \centering
    \caption{Relative wMSE computed over the testing set (score $\mathcal{S}$).}
    \label{tab:offline-da-results}
    \sisetup{round-mode=places, round-precision=3}
    \begin{tabular}{lrrrr}
    \headrow
    \thead{Correction mode} & \thead{$\mathcal{S}_{\mathsf{lnsp}}$} & \thead{$\mathcal{S}_{\mathsf{t}}$} & \thead{$\mathcal{S}_{\mathsf{vo}}$} & \thead{$\mathcal{S}_{\mathsf{d}}$} \\
    Zero correction & \num{1} & \num{1} & \num{1} & \num{1} \\
    Prediction & \num{0.7586344480514526} & \num{0.7544838786125183} & \num{0.8976179361343384} & \num{0.9187867641448975} \\
    Post-processing & \num{0.7490707039833069} & \num{0.7600661516189575} & \num{0.8763487339019775} & \num{0.9058694243431091} \\
    BL2020 (Post-processing) & \num{0.8799383640289307} & \num{0.981783390045166} & \num{0.9351458549499512} & \num{0.9498934149742126} \\
    \hline
    \end{tabular}
\end{table}

The results are reported in \cref{tab:offline-da-results}. For comparison, we also reported the scores of the neural network trained by \citet{bonavita-2020}, hereafter BL2020. For the BL2020 network, the scores are different from those reported by BL2020, which is primarily explained by the following three factors: (i) our score is computed in the non-normalised grid-point space (i.e. using $\boldsymbol{\mathcal{G}}$) whereas BL2020 computed their score in the normalised grid-point space (i.e. using $\widehat{\boldsymbol{\mathcal{G}}}$); (ii) we include data from all four seasons in our test set whereas BL2020 mainly included summer in their test set (the importance of this point is illustrated in \cref{sapp:offline-temporal-scores}); and (iii) our test data is at resolution T15 whereas BL2020 test data was at resolution T21. There are other sources of discrepancies (full Gauss--Legendre grid versus reduced Gauss--Legendre grid, test data mainly over 2021 versus test data over spring 2019, relative wMSE versus $R^2$ score, etc.) but we have checked that they only have a minor effect on the scores.

Despite these differences, our results confirm the findings of BL2020: the increments for lnsp and t are significantly more predictable than for vo and d. We can make two additional observations. First, the scores in prediction and in post-processing modes seem to be roughly similar, with a small advantage for post-processing (except for temperature). Second, the scores are significantly better for our trained neural networks than for the BL2020 trained neural network, which can be explained by two factors: our training set includes much more data ($\num{2740}$ state snapshots versus only $\num{243}$) and our neural network is much larger ($\num{1214876}$ parameters versus only $\num{380188}$ parameters).

\add{Finally, before presenting online training in the next section, note that additional diagnostics for offline training are illustrated in \cref{app:offline-additional-diagnostics}.}

\section{Step 2: online training}
\label{sec:online}

Now that the neural network for model error correction has been built and pre-trained offline, it is time to apply it online within data assimilation experiments. In \cref{sec:online-setup}, we briefly describe our data assimilation setup. Then, in \cref{ssec:offline_evaluation} we compare the neural network trained offline in post-processing and prediction modes using the online setup but with fixed parameters. Subsequently we will demonstrate the impact of online training in subsection \ref{ssec:online_training}. Finally, in subsection \ref{ssec:online_training_from_scratch}, we will investigate whether the NN 4D-Var can be adopted to train the neural network from scratch, bypassing the offline pre-training step.

\subsection{Data assimilation setup}
\label{sec:online-setup}

All the necessary algorithmic developments for NN 4D-Var were already accessible in OOPS, thanks to our prior work with the quasi-geostrophic model \citep{farchi-2023}. Consequently, to conduct our experiments, we only needed to create a model-specific implementation of an interface class for the neural network, acting as a bridge between the IFS and the neural network.

As explained in \cref{ssec:offline-nn-architecture}, the neural network is applied in grid-point space and the implementation in the IFS takes the following steps:
\begin{enumerate}
    \item the input fields are obtained via an inverse spectral transform of the IFS spectral t, lnsp, vo, and d fields, and then normalised as described in \cref{ssec:offline-nn-architecture};
    \item the neural network is then applied in the normalised, grid-point space, and the output is denormalised;
    \item the obtained correction is rescaled to the time step of the model, as explained in \citet{farchi-2023};
    \item finally, the rescaled correction is transformed into spectral space via a forward spectral transform to get the forcing term that is applied in the forecast following \cref{eq:methodology-nn4dvar-model}.
\end{enumerate}
The latter step uses the same infrastructure as the forcing formulation of weak constraint 4D-Var developed by \citet{laloyaux-2020}. 

All our experiments described in sub-sections to follow were performed using a standard research configuration of the weak constraint 4D-Var system with a $\qty{12}{\hour}$ assimilation window and using the latest available IFS cycle 48r1. The setup comprised three outer loops, with the model forecast resolution of TCo399\footnote{TCo$N$ means here that the data is at spectral resolution T$N$, interpolated onto a octahedral reduced Gaussian grid.} and the inner loop resolutions of TL95, TL159 and TL255 \footnote{TL$N$ means here that the data is at spectral resolution T$N$, interpolated onto a linear reduced Gaussian grid.}. In all our experiments employing the neural network model error correction we \delete{also }applied the corrections \add{in the assimilation, in all three outer itertions, and also }in the medium range ($\qty{10}{\day}$) forecasts. The corrections applied in the forecasts were updated every $\qty{12}{\hour}$. We chose the summer of 2022 (June, July, August), which is outside of the offline training data set, as the evaluation period and the standard weak constraint configuration of \citet{laloyaux-2020b} as the evaluation reference. \add{As illustrated in \cref{sapp:offline-temporal-scores}, summer is the season where the NN is most accurate offline. Therefore, one should keep in mind that the results might be not as good as the one presented below when evaluating in other seasons.}

\subsection{Comparison of post-processing and prediction mode in online experiments}
\label{ssec:offline_evaluation}

\begin{figure}[tbp]
    \centering
    \includegraphics[width=\linewidth]{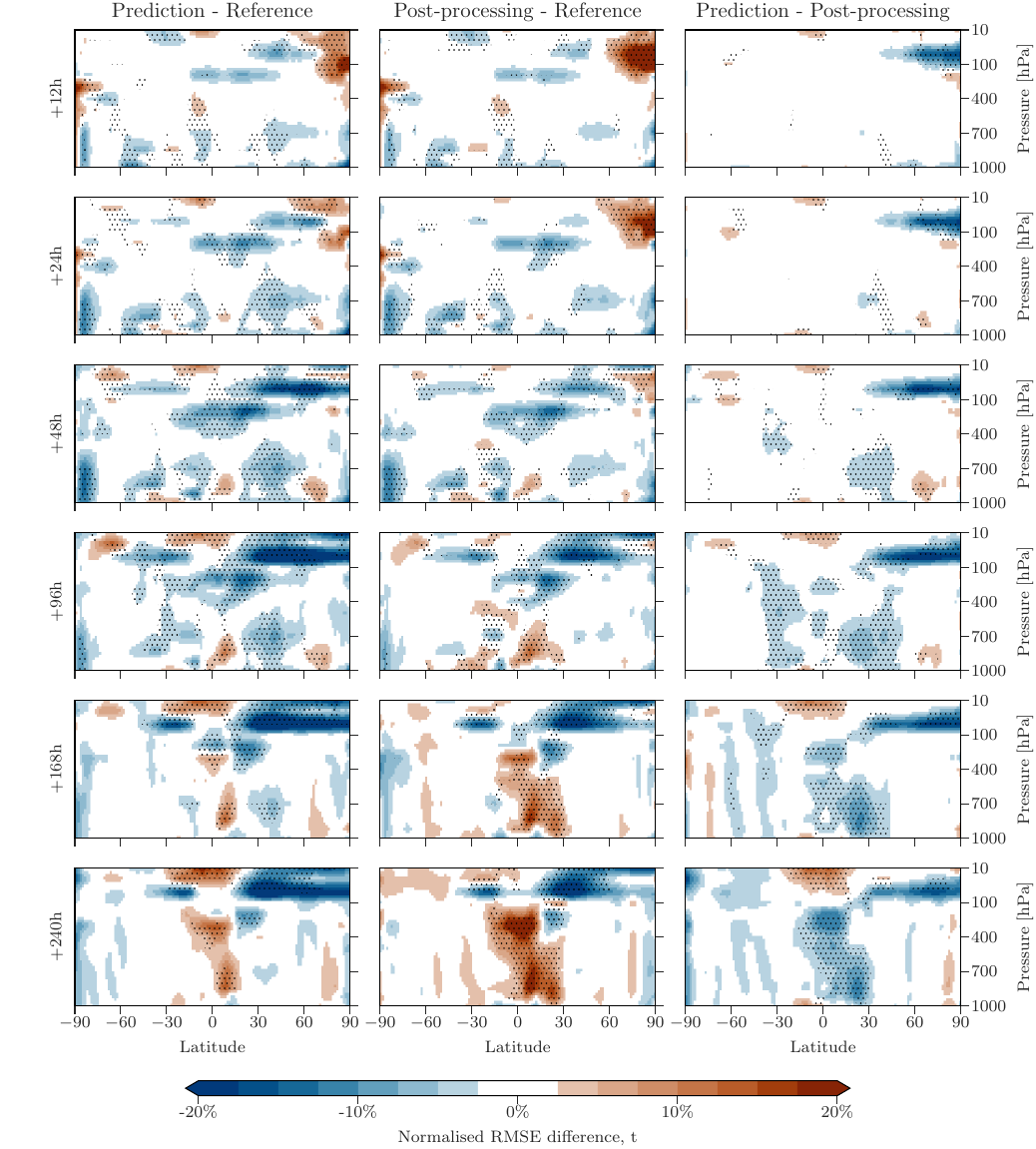}
    \caption{Change in normalised RMSE for the temperature field as a function of pressure level and latitude for forecast lead times ranging from $\qtyrange{12}{240}{\hour}$. The left panels compare strong constraint 4D-Var with the hybrid model in prediction mode \add{(Prediction)} to the standard weak constraint 4D-Var \add{(Reference)}. The middle panels compare strong constraint 4D-Var with the hybrid model in post-processing mode \add{(Post-processing)} to the standard weak constraint 4D-Var \add{(Reference)}. Finally the right panels compare strong constraint 4D-Var with the hybrid model in prediction mode \add{(Prediction)} to post-processing mode \add{(Post-processing)}. Blue colour indicates reduced errors and black dots marks statistically significant results using a $\qty{95}{\percent}$ two-sided t-test with a $\qty{25}{\percent}$ inflation of the confidence interval and \v{S}id\'{a}k correction for $\num{20}$ independent tests as recommended by \citet{geer-2016}.}
    \label{fig:online_pp_vs_pred_T_0001}
\end{figure}

While, as discussed in \cref{ssec:offline_training}, the choice between the post-processing and prediction mode of the neural network model of model error had little impact on the offline scores, we revisited this choice in the online experiments. We evaluated both pre-trained neural networks in 4D-Var experiments, keeping their parameters fixed, in other words using strong-constraint 4D-Var. \Cref{fig:online_pp_vs_pred_T_0001} shows the change in root mean-squared errors (RMSE) for temperature with respect to the standard weak constraint configuration as a function of latitude and pressure level for lead times ranging from $\qtyrange{12}{240}{\hour}$ when verified against the operational analysis (which is computed at a higher resolution using more outer loops, and hence can be considered a better estimate of the truth in our experiments). Both networks show significantly reduced errors above $\qty{100}{\hecto\pascal}$, particularly at long lead time. Below $\qty{100}{\hecto\pascal}$, there are mixed results. With the post-processing mode, the performance in the tropics is degraded especially with increasing lead time while no improvements in the extra tropics are visible beyond the first $\qty{72}{\hour}$. With the prediction mode, the degradation in the tropics at long lead time is less important and reduced errors are visible in the extra-tropics at all lead times.

\begin{figure}[tbp]
    \centering
    \includegraphics[width=\linewidth]{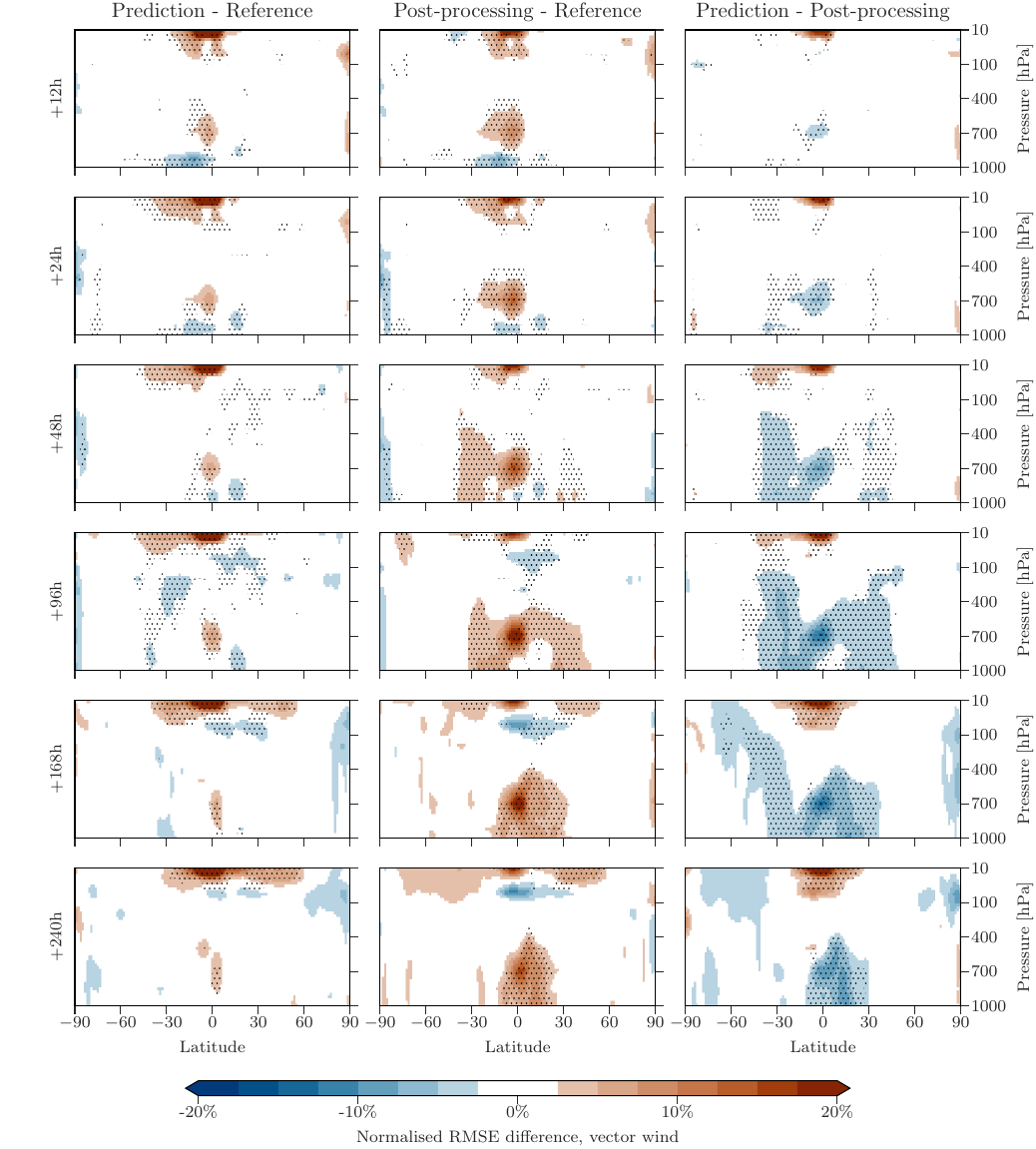}
    \caption{Same as \cref{fig:online_pp_vs_pred_T_0001} for the vector wind fields.}
    \label{fig:online_pp_vs_pred_VW_0001}
\end{figure}

The difference in performance between prediction and post-processing mode is even more evident for vector wind fields as can be seen in \cref{fig:online_pp_vs_pred_VW_0001}. Using the post-processing mode results in degradations both in the tropics and extra tropics for all model levels. The situation is different when using the prediction mode. While a negative signal in the stratospheric tropical region is visible at all lead times, the signal in the troposphere is much less negative and even positive in some situations.

\begin{figure}[tbp]
    \centering
    \includegraphics[width=0.5\linewidth]{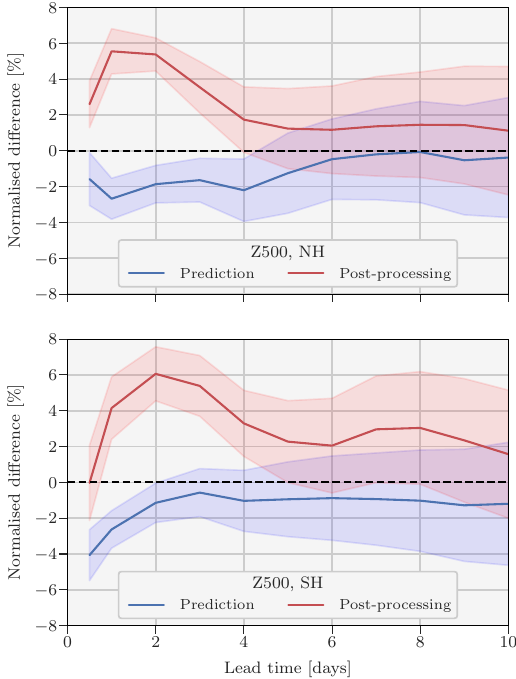}
    \caption{Change in normalised RMSE for geopotential at $\qty{500}{\hecto\pascal}$ in southern (bottom) and northern (top) hemisphere when using the hybrid model with the neural network model error component in post-processing (red) and prediction (blue) mode against the standard weak constraint formulation verified against own analysis. The shadow areas indicate the $\qty{95}{\percent}$ confidence intervals inflated by a $\qty{25}{\percent}$ factor and \v{S}id\'{a}k correction for $\num{8}$ independent tests as recommended by \citet{geer-2016}.}
    \label{fig:online_pp_vs_pred_Z_own_analysis}
\end{figure}

We conclude the evaluation of the hybrid model employing the neural network model error correction in post-processing and prediction mode by focusing on the evaluation of the normalised RMS error for the geopotential at $\qty{500}{\hecto\pascal}$ in the northern and southern hemispheres, representative of synoptic scale errors, in \cref{fig:online_pp_vs_pred_Z_own_analysis}. Applying the post-processing mode neural network results in degrading the geopotential in both hemispheres, while applying the prediction mode neural network results in a RMSE reduction of the order of $\qtyrange{1}{2}{\percent}$. Given the evidence of the superior performance when using the prediction mode neural network within the hybrid model described above, we stick with this choice when evaluating the impact of further online training in the following experiments.

\subsection{Online training from pre-trained network}
\label{ssec:online_training}

The NN 4D-Var formulation as defined in \cref{eq:methodology-nn4dvar-cost-gaussian} is now used to further train online, as part of the data assimilation process, the parameters of the neural networks which were pre-trained offline as described in \cref{ssec:offline_training}. We focus here only on the neural network model of model error pre-trained in the prediction mode. Extending the control vector of 4D-Var which \edit{hold}{holds} the state variables to include the parameters of the neural network necessitated specifying their background error model. We adopted the simple diagonal background error covariance matrix model $\mathbf{P}=p^{2}\mathbf{I}$ described in \cref{ssec:two-step-training}. In the absence of a good estimate of the statistics of the background errors of the neural network parameters, we performed a sensitivity study choosing a constant value for $p$, the parameter error standard deviation, between $\num{0.001}$, $\num{0.0005}$ and $\num{0.0001}$. We only show the results for what we found to be an optimal choice of $p=\num{0.0005}$ among the tested values with the remaining choices showing signs of either \edit{over fitting}{overfitting} (for $p=\num{0.001}$) or limited impact (for $p=\num{0.0001}$) with respect to the pre-trained neural network. \add{In the first case, the neural network parameter updates are large and reflect model error patterns that are specific to the current window and hence do not generalise well to subsequent windows. In the second case, the parameter updates are so small that there is no significant evolution of the neural network throughout the experiment. Ideally, we would like to have more control over the parameter updates, for example by progressively decreasing the value of $p$ as used by \citet{farchi-2021b} for low-order models, but this would involve too much tuning in computationally extensive experiments.}

Before discussing the results, it is worthwhile to remark that we saw no evidence suggesting that extending the control vector to include the parameters of the neural network (in practice, the $\num{1214876}$ parameters represent less than $\qty{2}{\percent}$ of the the control vector at each of the three inner loops) had any impact on the convergence of NN 4D-Var. For all the investigated choices of $p$ the mean number of inner loop iterations across all outer loops averaged over all cycles of an experiment was barely different from that of the standard weak constraint configuration. Overall, the NN 4D-Var showed neutral impact on the computational performance of the ECMWF assimilation system for the considered configuration.

\begin{figure}
    \centering
    \includegraphics[width=0.7\linewidth]{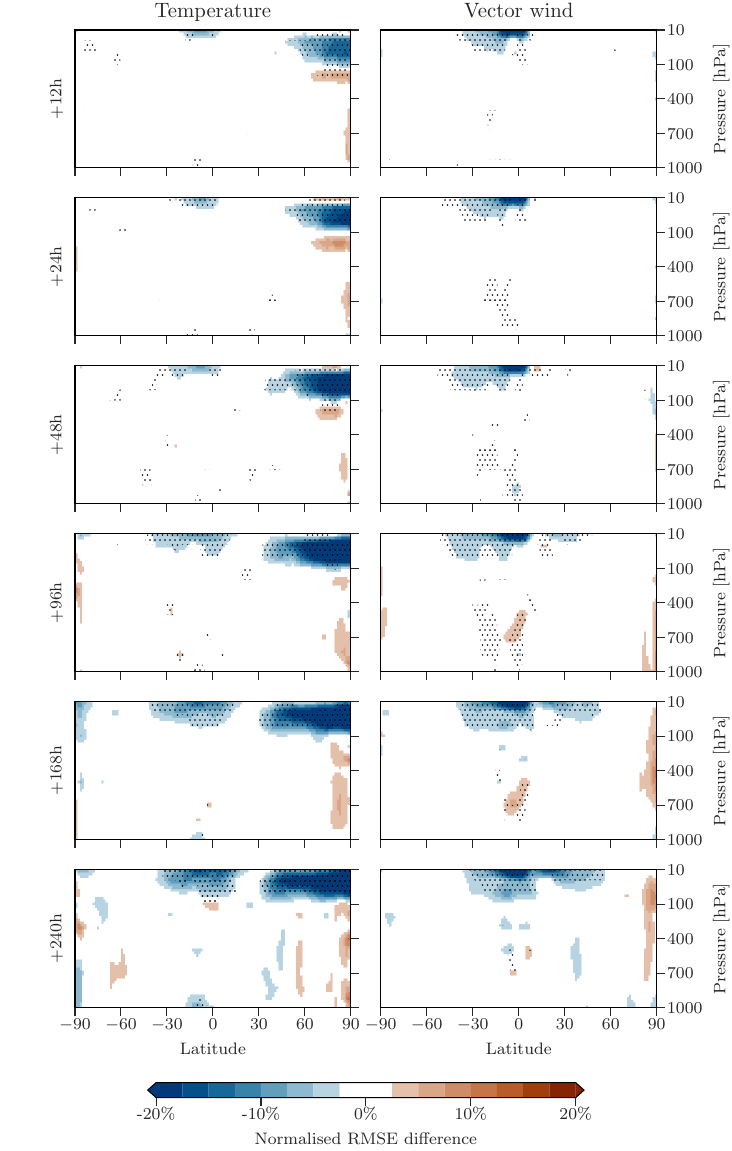}
    \caption{Same as \cref{fig:online_pp_vs_pred_T_0001} comparing NN 4D-Var to strong constraint 4D-Var, both with the hybrid model, i.e. online training of the neural network compared to offline pre-trained constant neural network, for the temperature (left panels) and vector wind (right panels) fields.}
    \label{fig:online_vs_offline_T_VW_ifs_0001}
\end{figure}

It is expected that allowing the NN 4D-Var to further adjust the parameters of the neural network model of model error as part of the data assimilation process should result in improved forecast scores following the results of \citet{farchi-2023}. \Cref{fig:online_vs_offline_T_VW_ifs_0001} shows the normalised change in forecast RMSE for the temperature and vector wind fields when performing online training of the neural network parameters compared to when using only the pre-trained neural network parameters, both verified against the operational analysis. The online training allowed to significantly reduce the temperature errors in the stratosphere at all lead times. The improvements are most significant in the northern hemisphere with up to $\qty{30}{\percent}$ reduction of RMSE above $\qty{100}{\hecto\pascal}$. Almost no impact is visible below $\qty{100}{\hecto\pascal}$. The vector wind field RMSE is also significantly reduced above $\qty{100}{\hecto\pascal}$. Recalling the right panel of \cref{fig:online_pp_vs_pred_VW_0001}, it is precisely where the pre-trained network does not perform well compared to the standard weak constraint configuration. A hint of degradation is visible in the troposphere in the tropics, in particular at longer forecast lead times.

\begin{figure}[tbp]
    \centering
    \begin{tabular}{ccc}
        \footnotesize
        Online - Reference & \footnotesize Online - Offline &\footnotesize Online from scratch - Reference \\
        \fbox{\includegraphics[height=0.45\textheight]{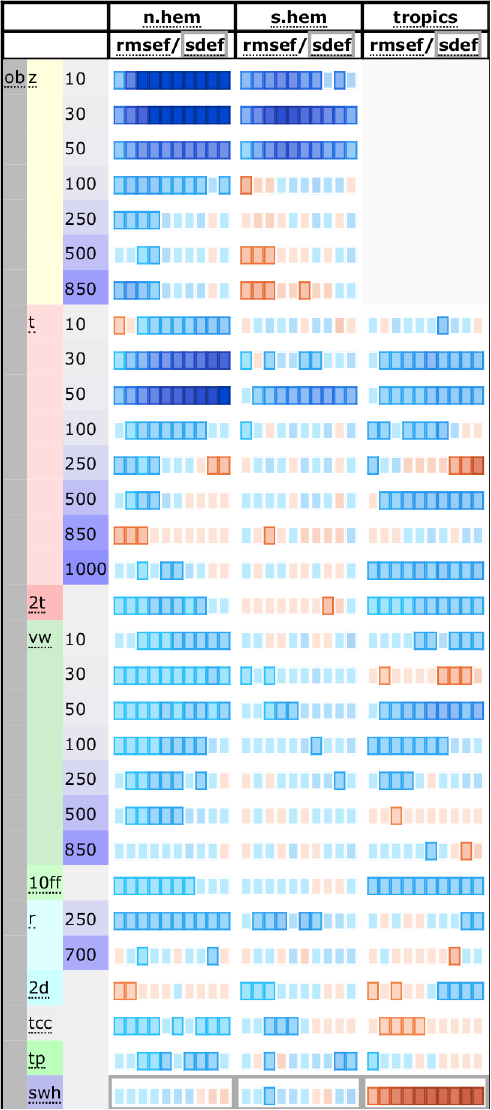}} & \fbox{\includegraphics[height=0.45\textheight]{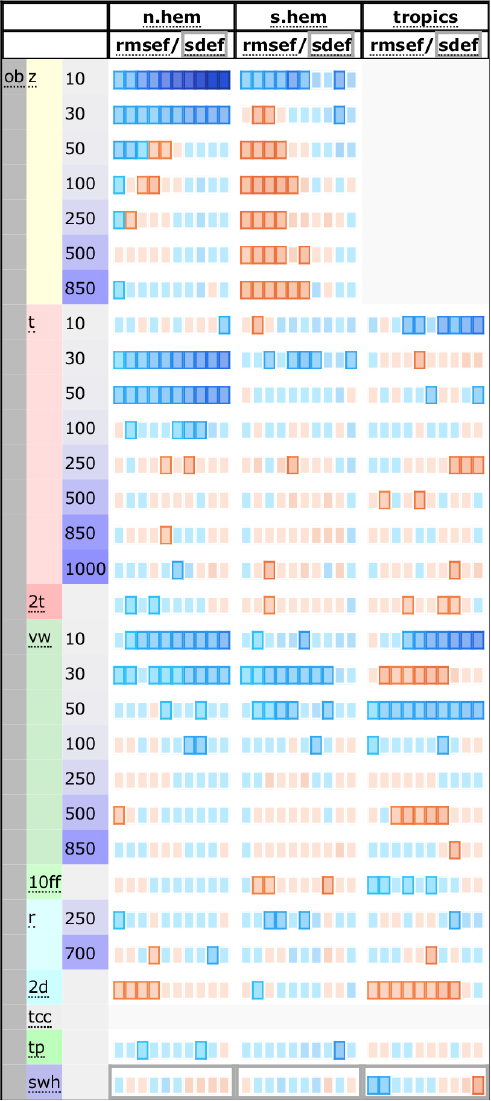}} & \fbox{\includegraphics[height=0.45\textheight]{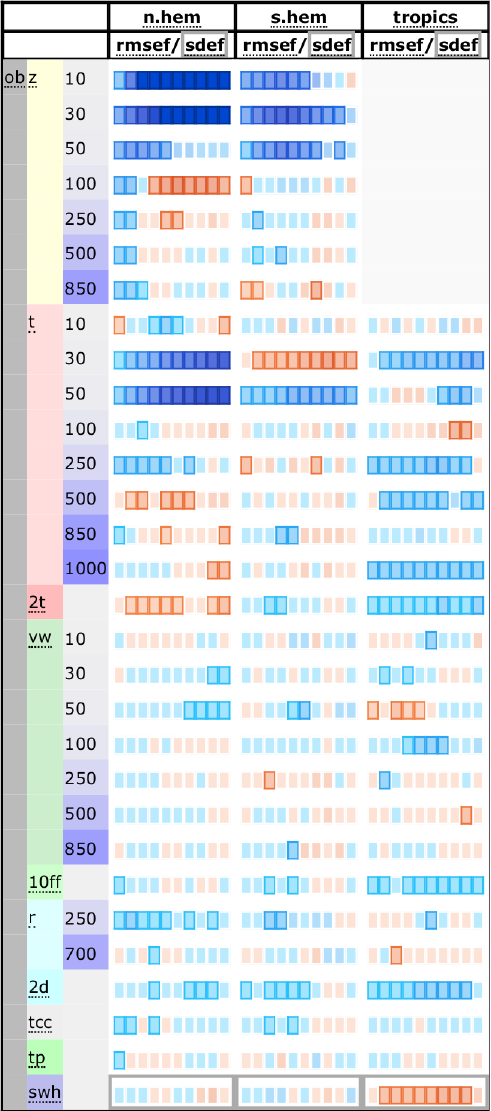}}
    \end{tabular}
    \caption{Score cards showing the change in forecast RMSE verified against independent observations for geopotential (z), temperature (t), two-meter temperature (2t), vector winds (vw), ten-meter wind speed (10ff), relative humidity (r), two-meter dew point (2d), total cloud cover (tcc), total precipitation (tp), and significant wave height (swh) fields as a function of pressure level (for three-dimensional fields). Left panel: the impact of online training \add{(Online)} with respect to the standard weak constraint configuration \add{(Reference)}. Middle panel: the impact of online training \add{(Online)} with respect to offline training \add{(Offline)}. Right panel: the impact of online training from scratch (i.e. without offline pre-training, \add{Online from scratch}) with respect to the standard weak constraint configuration \add{(Reference)}. The horizontal bars represent lead times spanning from 1 to 10 days. The blue and red colour and their intensity reflect the reduction and degradation of the forecast skill, respectively.}
    \label{fig:scorecards}
\end{figure}

While the verification against the operational analysis can be considered to provide a good first glimpse of the impact of online training on the forecast performance, the ultimate assessment is carried out against independent observations. The left panel of \cref{fig:scorecards} shows a forecast RMSE scorecard demonstrating the performance of a hybrid model with online training of the neural network parameters within the NN 4D-Var framework compared to the standard weak constraint formulation of \citet{laloyaux-2020b}. The right panel of this figure shows the impact of online training with respect to when using only a pre-trained neural network in the hybrid model. In both cases the change in forecast scores is verified against observations.

Considering first the left panel in \cref{fig:scorecards}, what is not evident from the verification against the operational analysis, is that the positive impact of the NN 4D-Var stretches throughout the whole atmospheric column for both temperature and vector wind fields, in particular in the northern hemisphere and the tropics. Overall, the impact on forecast RMSE of all variables is positive in the northern hemisphere and tropics, while it is relatively modest in the southern hemisphere with the exception of the stratosphere. Interestingly, a significant, positive impact is also visible for variables that were not explicitly corrected by the neural network, namely for the relative humidity (r), total cloud cover (tcc) and total precipitation (tp) fields in extra tropics. It is also worth noting the improved two-meter temperature (2t) scores in the northern hemisphere (of the order of $\qty{1}{\percent}$) and in the tropics (up to $\qty{2}{\percent}$), which are of practical relevance to forecast users. On the downside, the temperature scores at $\qty{850}{\hecto\pascal}$ are slightly degraded at short lead times, in particular in the northern hemisphere. 

The middle panel in \cref{fig:scorecards} provides further evidence of a positive impact of online training of the neural network within NN 4D-Var. The picture is globally similar to that of \cref{fig:online_vs_offline_T_VW_ifs_0001} showing the impact on forecast RMSE for temperature and vector wind fields verified against the operational analysis. Apart from a large positive impact in the stratosphere, the results point to a small degradation in the geopotential in southern hemisphere at short lead times. The fact that the online training improves the most the stratosphere can be explained by the fact that this is where the model has the most prominent large scale biases, which evolve in a slow and predictable fashion as highlighted by \citet{laloyaux-2020b}.

\begin{figure}
    \centering
    \includegraphics[width=\linewidth]{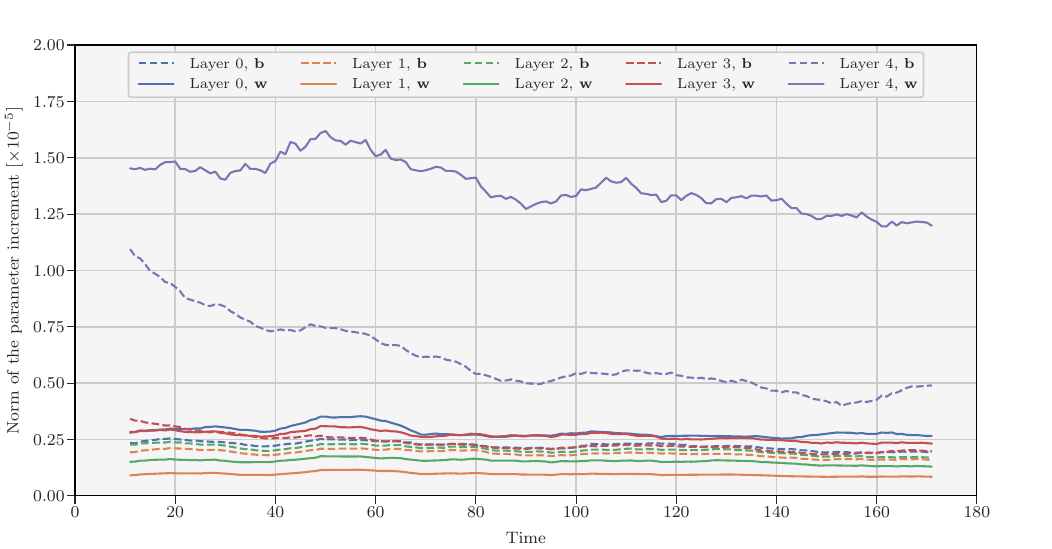}
    \caption{\add{Evolution of the norm of the parameter increments as a function of time throughout the $\num{181}$ assimilation windows of the online training experiment. To make the figure easier to read, we only show a $\qty{10}{\day}$ running-average of the values. The norm is computed independently for each layer and for each parameter type: bias ($\mathbf{b}$, dashed lines) or weights ($\mathbf{w}$, continuous lines). Layers $\num{0}$ to $\num{3}$ (in blue, yellow, green, and red) are the four hidden layers, and layer $\num{4}$ (in purple) is the output layer.}}
    \label{fig:parameter-increment}
\end{figure}

\add{To conclude this section, let us discuss the evolution of the neural network parameters throughout the online training experiment. To this end, \cref{fig:parameter-increment} illustrates the norm of the parameter increment, defined here as}
\begin{equation}
    \left\| \delta \mathbf{p} \right\| \triangleq \sqrt{\frac{1}{N_{\mathsf{p}}}\sum_{i=1}^{N_{\mathsf{p}}}\delta p_i^2},
\end{equation}
\add{where $N_{\mathsf{p}}$ is the total number of parameters and $\delta \mathbf{p}$ is the parameter increment. First, it is clear that the parameter increments are very small, which means that the parameter do not evolve much over the 3 month of the experiment. This is confirmed by the evolution of individual parameters (not illustrated here). This was expected, because we chose a very small background error covariance matrix for model parameters $\mathbf{P}$ to avoid overfitting. Second, it is also clear that the parameters that evolve the most are those of the last layer, which could indicate that online learning mostly provides a fine-tuning of the neural network. Third, the norm of the parameter increment decreases over time, even though the background error covariance matrix for model parameters $\mathbf{P}$ has been kept constant. This implies that the neural network parameters are slowly reaching convergence. However, after three months of assimilation, the system has not reached a steady-state, which means that the results could be further improved with longer training.}

\subsection{Online training from scratch}
\label{ssec:online_training_from_scratch}

Impressed by the ability of the online training of the neural network within NN 4D-Var to rapidly and significantly improve the stratospheric forecast scores compared to the pre-trained network and over a relatively short three-month period, it is of interest to investigate if the NN 4D-Var data assimilation optimisation framework is able to effectively train the neural network from scratch, that is, when bypassing the offline pre-training step altogether. While the offline pre-training relies on bespoke non-linear optimisation methods based on the stochastic gradient descent algorithm developed for efficient training of deep neural networks, the incremental 4D-Var relies on an iterative Gauss-Newton optimisation technique with the Lanczos algorithm used for the minimisation of a sequence of convex quadratic cost functions. 

The right panel of \cref{fig:scorecards} shows a forecast RMSE score card verified against observations when training the neural network within NN 4D-Var framework without previous offline pre-training compared to the reference standard weak-constraint configuration. As can be seen, the network managed to learn the structure of systematic errors in the stratosphere with impressive improvements visible for the temperature and wind vector fields above $\qty{50}{\hecto\pascal}$. Interestingly the forecast RMSE are also reduced throughout the whole atmospheric column in the tropics. This is where the IFS model is known to have the large biases and the neural network is able to recognise them very quickly. As expected, the extra tropical tropospheric errors are less predictable and therefore harder to learn, yet one can already notice small (although not yet statistically significant) positive impact there too. \add{Finally, even though the results of online learning from scratch are positive, they are not as good as the results of online learning from the pre-trained network (left panel of \cref{fig:scorecards}). This confirms that offline training is a valuable pre-training step to speed-up online training.}

\section{Discussion}
\label{sec:discussion}

\subsection{Architecture of the neural network}
\label{ssec:conclusions-architecture}

The primary objective of the present work is to develop a neural network model error correction that can be used later on in data assimilation and forecast applications. 

For practical reasons, we have decided to use a vertical/column architecture, where the same neural network is applied independently to each atmospheric column. Switching to a non-column approach is possible, but would require more work to implement online due to the parallelisation aspects of the IFS. At this point, there is no practical evidence that a non-column approach would be significanlty more accurate for the model error correction task \citep[see in particular][for comparison]{chen-2022}. Furthermore, the major drawback of the column approach, the fact that horizontal spatial relationships are ignored by the neural network, can be circumvented using additional predictors, such as the horizontal (zonal and meridional) gradients of the input fields, as proposed by \citet{kochkov-2023}. We have tested this approach in our set of offline experiments, but decided not to include it in our set of online experiments, because the improvement was only marginal (of the order of a few percents in the offline scores). This is, however, likely to change with resolution, as we believe that at higher resolution the horizontal gradients would provide a valuable insight on the small-scale variation of the variables.

Beyond the choice of a column-based architecture, we have opted for a fully-connected neural network. This choice is computationally feasible because, in each vertical column, the number of input and output variables is reduced to, respectively, $\num{420}$ and $\num{412}$. However, one must keep in mind that fully-connected neural networks do not scale well and will not be a viable option with increased vertical resolution or with more predictors (e.g. more variables or simply the spatial gradients as suggested in the previous paragraph). On one hand, convolutional neural networks would be an attractive alternative, because the vertical layers are supposed to have only local influence on each other. On the other hand, the physical processes (and hence the model error) vary over model levels and, more importantly, model levels are not evenly distributed in the vertical direction. The first issue could be easily solved by adding altitude or pressure coordinate as extra predictor to the neural network, but the second issue requires more attention. A possibility could be to use locally connected layers in place of convolutional layers, with the caveat that locally connected layers are usually heavier in terms of parameters than convolutional layers. Another possibility could be to use an encoder-decoder architecture, mapping from the original space (where model levels are not evenly distributed) to a latent space. In addition, it is also very much possible that using standard machine learning \enquote{tricks} such as residual connections would improve the offline performance.

%\add{[This paragraph has moved from the following section to the present section]} 
In summary, there is certainly room for improvement in the design of the neural network architecture. Nevertheless, one has to keep in mind that the ultimate goal is to have the model error correction included online, in data assimilation and forecast experiments. This means that the main performance criterion should be the online scores, and consequently, that the neural networks should be implemented online, in \edit{Fortran}{the same programming language as the physical model, which is Fortran for the IFS}. In our experiments, this was made possible by using the Fortran neural network library \citep[FNN, ][]{fnn-2022}, which supports fully-connected neural networks. While the FNN library can be easily extended to convolutional layers, specific architectures, in particular with residual connections, will remain difficult to implement without manually coding them.

\subsection{Discrepancy between offline and online scores}

Throughout the offline and online experiments, the performance of the neural network has been illustrated in different conditions. The overall justification for using the two-step training, first offline then online, relies on the idea that the offline experiment does provide a valuable pre-training of the neural network, since it can make use of a \enquote{large} dataset (more than three years offline versus only three months online). Our experiments confirm, to a certain degree, that offline pre-training is useful, but they also show that there is a significant discrepancy between offline and online scores. In particular, the offline errors are lowest near the surface, while the online errors are lowest in the upper levels. The difference is coming mainly from two factors: (i) the IFS version is different in the offline and online experiments, and (ii) in the offline experiments, the interaction between the IFS and the model error correction is neglected. From our experience, the first factor has a smaller impact compared to the second factor. Taking into account the interactions between the IFS and the model error correction in the offline experiments is possible, provided that an auto-differentiable version of the IFS is available, e.g. using an emulator. Such an emulator could be based, for example, on one of the latest MLWP models, fine-tuned to the latest IFS model version.

\subsection{Implementing the time-dependent correction}

For the new 4D-Var variant, NN 4D-Var, the assumption of constant model error over the window is not fundamentally required: it is used only to make NN 4D-Var closer to the existing weak-constraint 4D-Var and hence to reduce the initial implementation burden of the method. Yet, we expect model errors to be time-dependent within a window, and hence it is desirable to remove the assumption of constant model error over the window. In practice, this means additional implementation work, but we believe that it is not beyond reach. However, in that case the experimental protocol will most probably have to be reworked.

Within the current setup, during the offline pre-training step, the network is exposed to analyses and analysis increments which are always located at the start of the operational data assimilation window, that is 9:00 UTC or 21:00 UTC. On one hand, it is always possible to use a neural network that has been trained in this way and hope that online training will be sufficient to make the neural network able to predict the daily variability of model error. On the other hand, there are ways to improve offline pre-training in this context. For example, it is possible to augment the training dataset by including the analyses and analysis increments within the window (and not only at the start of the window). However, we must keep in mind that, rigorously speaking, the analysis increments are a proxy for model error only at the start of the window.

\subsection{Towards higher resolution in offline learning}

In our offline experiments, the dataset has been truncated to a very coarse T15 resolution. This choice was made for practical reasons, but also because we expect model errors to be prominent at large scales \citep{laloyaux-2020b}. Nevertheless, we have shown that using higher resolution training data can increase the accuracy, especially in a multivariate setup. Our assumption is that this is coming from the fact that some variables are driven by large scales (t and lnsp) while other are driven by smaller scales (vo and d). At this point, it is still unclear (i) how much resolution is needed to get an accurate representation of the analysis increments (ii) what is the finest predictable resolution, especially taking into account the size of our training dataset (about four years). Beyond these questions, further research is also needed to determine what is the best strategy for going from offline to online. In particular, whether we should keep the same resolution in offline and online experiments or whether we can actually increase the resolution remains an open question.

\subsection{Added value of online learning}

First of all, a major benefit of our developments is to provide a framework where a model error correction can be evaluated in close to operational conditions, i.e. with data assimilation cycles followed by medium-range forecasts. In particular, this allowed us to objectively compare prediction and post-processing mode and conclude, as we expected, that the prediction mode is better suited to online experiments.

Second, our experiments clearly show that online learning is effective and does improve the network beyond offline pre-training, which confirms the conclusions of \citet{farchi-2021b, farchi-2023} with low order models. An important part of the improvement is coming from the fact that the developed online framework is able to take into account the interactions between the IFS and the model error correction throughout the assimilation window. However, we believe that another significant part of the improvement is coming from the fact that online learning directly targets the observations, whereas offline learning targets the analysis increments. 

Finally, our online learning framework, NN 4D-Var, can be seen as a natural extension or reformulation of weak-constraint 4D-Var, a well established data assimilation method. Consequently it is built around the concept of joint learning of model state and model parameters. Alternatively, it is possible to solely focus on model parameter estimation, for example by removing the initial model state from the control variable of the NN 4D-Var cost function defined in \cref{eq:methodology-nn4dvar-cost-gaussian} and replacing it by a fixed, reference analysis. Removing the need for cycling of the atmospheric analysis would facilitate the improvement of the efficiency of the neural network training by allowing to evaluate a batch of data assimilation cycles in parallel. Furthermore, the predictive skill of the neural network could also be improved by extending the data assimilation window beyond the standard 12h. This would allow the observations to better constrain the neural network over longer forecast lead times, which has been proven useful for many MLWP \citep[e.g.,][]{lam-2023, kochkov-2023}

\section{Conclusions}
\label{sec:conclusions}

This work is a step forward in the direction of developing a hybrid system, where a physics-based model (namely the IFS) is supplemented by a neural network, for operational data assimilation and forecasting applications. In practice, the neural network can be seen as a model of model error of the physics-based model. For practical reason, we choose a fully-connected column neural network. This neural network is trained in a two-step process: first offline then online. 

In the offline training step, the neural network is trained to predict the analysis increments, which can be seen as a proxy for the model error developing over one data assimilation window. The analysis increments are extracted from a dataset gathering the operational analyses and background forecasts produced by ECMWF between 2017 and 2021 at T15 resolution and interpolated on a regular Gaussian grid. Within this dataset, the trained neural network is able to predict $\qtyrange{10}{25}{\percent}$ of the increments depending on the atmospheric variables.

Once trained offline, the neural network is plugged into the IFS, thanks to the FNN library, and hence can be used online in data assimilation and forecast experiments. Starting with regular data assimilation experiments, where only the model state is estimated (i.e. the neural network parameters are not estimated), we show that the neural network correction is effective, which translates into reduced forecast errors in many conditions, for example we observe an RMSE reduction of the order of $\qtyrange{1}{2}{\percent}$ for the geopotential at $\qty{500}{\hecto\pascal}$. The network is then further trained online, using the new 4D-Var variant, NN 4D-Var. The accuracy improvements are then reflected in the scorecards, with reduced forecast errors in almost all conditions. We conclude that NN 4D-Var can be considered as an effective online training tool for neural network based model error corrections.

Many possibilities are open for future work. Focusing on the offline pre-training step, we have seen that, in the multivariate setup, increasing the resolution of the training data effectively increases the accuracy of the neural network. Even though only large scale model error is assumed to be predictable, the T15 resolution selected in the \delete{our }experiments is probably insufficient, especially for vorticity and divergence. When it comes to the online experiments, we believe that one of the most promising \edit{perspective}{perspectives} is to extend the 4D-Var formulation to a time-dependent correction within each window. This will require additional implementation work, but would enable the neural network to represent the daily variability of model error.

\section*{Acknowledgements}
CEREA is a member of Institut Pierre--Simon Laplace. \add{The authors thank two anonymous reviewers whose comments and suggestions helped improving the manuscript.}

\section*{Conflict of interest}
There is no conflict of interests.

\section*{Data availability statement}
The offline training dataset described in \cref{sec:offline} is available with the ECMWF MARS archive (https://apps.ecmwf.int/mars‐ catalogue/), registration required. The results of the online experiments described in \cref{sec:online} are available on demand.

\appendix
\section{Sensitivity to the size of the dataset in offline training}
\label{app:offline-size-sensitivity}

\begin{figure}[tbp]
    \centering
    \includegraphics[width=\linewidth]{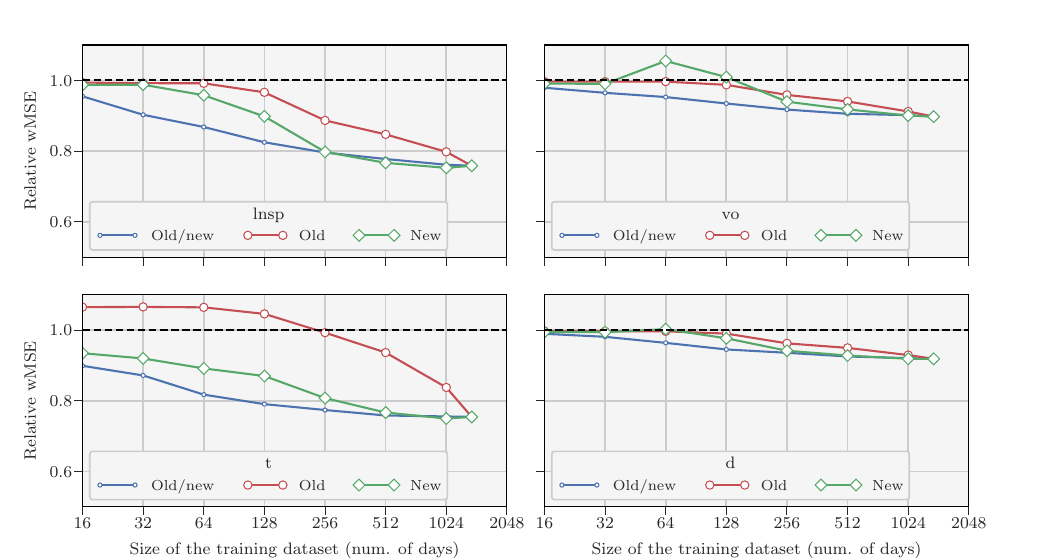}
    \caption{Relative wMSE for lnsp (top left panel), t (bottom left panel), vo (top right panel), and d (bottom right panel), as a function of the size of the training dataset in the \enquote{old and new} (in blue), the \enquote{old} (in red), and the \enquote{new} (in green) strategy.}
    \label{fig:offline-size}
\end{figure}

\begin{figure}[tbp]
    \centering
    \includegraphics[width=\linewidth]{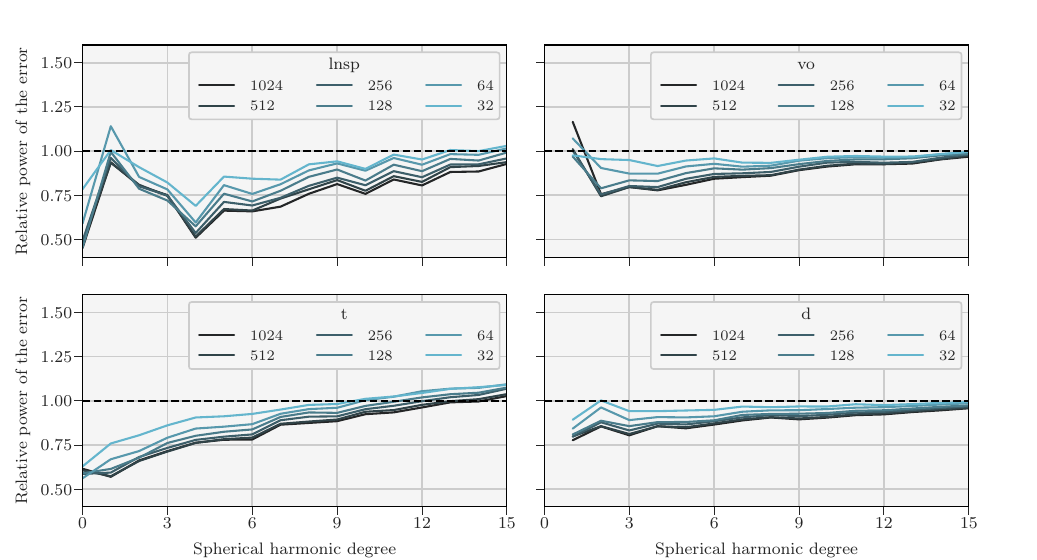}
    \caption{Relative averaged power spectra of the prediction errors for lnsp (top left panel), t (bottom left panel), vo (top right panel), and d (bottom right panel) when the size of the dataset is progressively reduced from $\qty{1024}{\day}$ (in black) to $\qty{32}{\day}$ (in teal).}
    \label{fig:offline-size-spec}
\end{figure}

In this \edit{section}{appendix}, we investigate the sensitivity of the \add{offline} accuracy of the network as a function of the size of the training dataset. To do so, we take the \edit{reference}{training} setup of \cref{sec:offline}, and we progressively reduce the number of days in the training dataset, using three different strategies. In the first strategy, \enquote{old and new}, the selected days are equally distributed over the entire available data. In the second strategy, \enquote{old}, the selected days are the most ancient days in the available data. Finally, in the third and last strategy, \enquote{new}, the selected days are the most recent days in the available data. In all cases, the exact same neural network is trained. The relative wMSE are shown in \cref{fig:offline-size}. In addition, we compute the relative averaged error power spectra in the \enquote{old and new} strategy and show the results in \cref{fig:offline-size-spec}.

Without surprise, the neural network gets more accurate when the dataset gets larger. Interestingly, increasing the size of the dataset reduces the errors of the network at all spectral degrees. Furthermore, the \enquote{new} strategy leads to lower errors in almost all cases than the \enquote{old}. This confirms that the older data are of lesser interest for our study, because of the continuous updates in the IFS which progressively changes the model and hence the model errors. Therefore, there is a balance to find between two competing effects: on the one hand including more data is beneficial to train large neural network, on the other hand the additional data is older and hence provide less information. In our experiments, the balance seems to be optimal for a dataset of $\qty{1024}{\day}$ in the \enquote{new} strategy. However, this number is likely to change depending on the size of the neural network.

\section{Additional diagnostics for offline training}
\label{app:offline-additional-diagnostics}

\add{[This section has moved to the appendix]} To extend the analysis \add{of the offline training results}, we computed the relative wMSE on limited parts of the testing set as well as several other diagnostics. The results are presented in the following subsections for the prediction mode only. We have checked that the results for the post-processing mode are qualitatively and quantitatively similar.

\subsection{Temporal scores}
\label{sapp:offline-temporal-scores}

\begin{figure}[tbp]
    \centering
    \includegraphics[width=\linewidth]{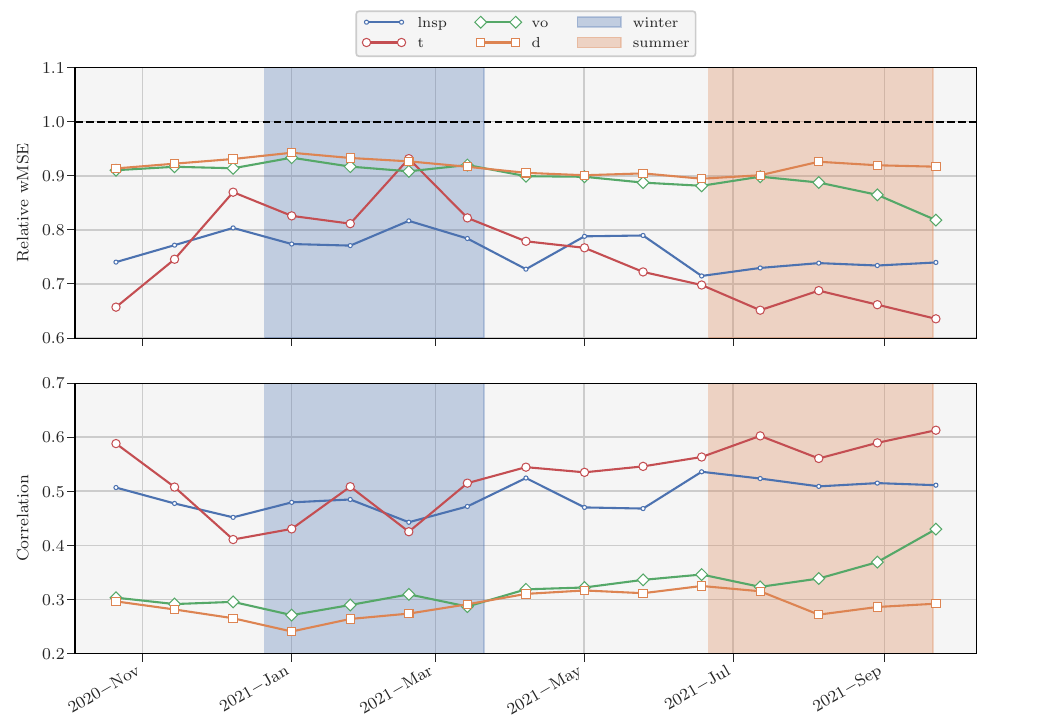}
    \caption{Averaged temporal relative wMSE (top panel) and averaged Pearson correlation over space (bottom panel) for each of the four variables: lnsp in blue, t in red, vo in green, and d in yellow. Winter 2020/2021 is highlighted in blue and summer 2021 in yellow.}
    \label{fig:offline-temporal-scores}
\end{figure}

Let us start by computing the temporal relative wMSE, in other words the relative wMSE computed independently for each state snapshot (one value for each variable and each $t\in\mathcal{T}_{\mathsf{test}}$). In addition, we compute the Pearson correlation over space between the predicted and the actual increments (again one value for each variable and each $t\in\mathcal{T}_{\mathsf{test}}$). The results are averaged over each batch of consecutive training days, and then illustrated in \cref{fig:offline-temporal-scores}.

Visually, there is more variability in the scores for t and lnsp than for vo and d. For t, the neural network is significantly more accurate in summer than in winter. This is also the case, although to a lesser extent, for the other variables. Our hypothesis is that winter errors are more connected to dynamical situations (e.g. misplaced frontal systems) while summer errors are more connected to systematic model deficiencies (e.g. depth of nighttime inversions). This underlines the importance of having a representation of a full year in the validation and testing datasets. In addition, the Pearson correlation over space is surprisingly higher than one could expect (from the relative wMSE values). This would tend to indicate that the neural network is unable to estimate the spatial mean and variance of the increments. In our case, we have checked that the squared bias contribution to the wMSE is always lower than $\qty{2}{\percent}$ and on averaged lower than $\qty{0.3}{\percent}$, meaning that the neural network is able to provide an accurate estimation of the spatial mean. By contrast, we have found that the neural network significantly underestimates the spatial variance of the increments (by a factor between $\num{1.5}$ and $\num{10}$ depending on the variable). This is a typical feature of deterministic neural networks trained with a point-wise objective such as the mean-squared error, which tend to smooth out predictions to circumvent the double penalty effect for patterns that are difficult (or impossible) to predict. 

\subsection{Spatial scores}
\label{sapp:offline-spatial-scores}

\begin{figure}[tbp]
    \centering
    \includegraphics[width=0.5\linewidth]{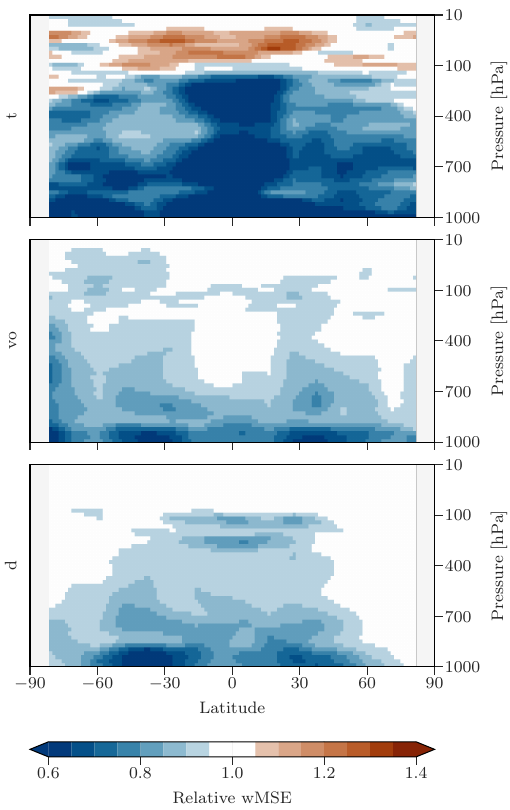}
    \caption{Slice of relative wMSE over latitude and model levels (in pressure coordinate) for each of the three atmospheric variables: t (top panel), vo (middle panel), and d (bottom panel).}
    \label{fig:offline-slice-mse}
\end{figure}

We now compute a spatial slice of relative wMSE over latitude and model levels, in other words the relative wMSE computed independently for each latitude node and each model level. The results are illustrated in \cref{fig:offline-slice-mse}.

Overall, it is clear that the neural network is most accurate close to the surface. For t, the neural network remains very accurate up until $\qty{100}{\hecto\pascal}$. For vo and d, the scores until $\qty{100}{\hecto\pascal}$ are still positive, although significantly less than at the surface. Conversely at higher altitude, between $\qtylist[list-units = single]{10;100}{\hecto\pascal}$, the estimation of model error significantly deteriorates. It even increases the errors for t. This degradation can most probably be attributed to the influence of weak constraint 4D-Var. Indeed, in our experiment the training set uses analyses and increments mostly produced with strong constraint 4D-Var, while the testing set relies on solely on weak constraint 4D-Var data. We conclude that weak constraint 4D-Var significantly alters the analysis increments between $\qtylist[list-units = single]{10;100}{\hecto\pascal}$ (where it is active) thereby undermining the assumption that the analysis increment serves as a reliable proxy for model error. Further investigations are necessary to address this challenge. One potential avenue involves training the neural network to predict the sum of the analysis increment and the weak constraint forcing, as opposed to solely the analysis increment, as currently implemented.

\subsection{Spectral analysis}
\label{sapp:offline-spectral-analysis}

To conclude this series of additional diagnostics, we compute the power spectra of (i) the neural network inputs, (ii) the expected outputs, (iii) the predictions, and (iv) the prediction errors (difference between the expected outputs and the predictions). For a field $\mathbf{x}\in\mathbb{R}^{272}$ in spectral space (at resolution T15 here), the power spectrum of $\mathbf{x}$ at spectral degree $l\leq15$ is defined by
\begin{equation}
    \mathcal{P}_{l}\left(\mathbf{x}\right) \triangleq \sum_{c=0}^{1}\sum_{m=0}^{l} x^{2}_{c, l, m},
\end{equation}
where $x_{c, l, m}$ is the real (if $c=0$) or imaginary (if $c=1$) component of $\mathbf{x}$ corresponding to the spherical indices $\left(l, m\right)$. By construction, we have
\begin{equation}
    \left\|\mathbf{x}\right\|^{2} = \sum_{l=0}^{15} \mathcal{P}_{l}.
\end{equation}
Furthermore, the spatial mean and variance of $\mathbf{x}$ in grid-point space (i.e. over latitude and longitude) are given by
\begin{align}
    \mathrm{mean}\left(\mathbf{x}\right) &= \sqrt{\mathcal{P}_{0}},\\
    \mathrm{var}\left(\mathbf{x}\right) &= \left\|\mathbf{x}\right\|^{2}-\mathrm{mean}\left(\mathbf{x}\right)^{2} = \sum_{l=1}^{15} \mathcal{P}_{l},
\end{align}

\begin{figure}[tbp]
    \centering
    \includegraphics[width=\linewidth]{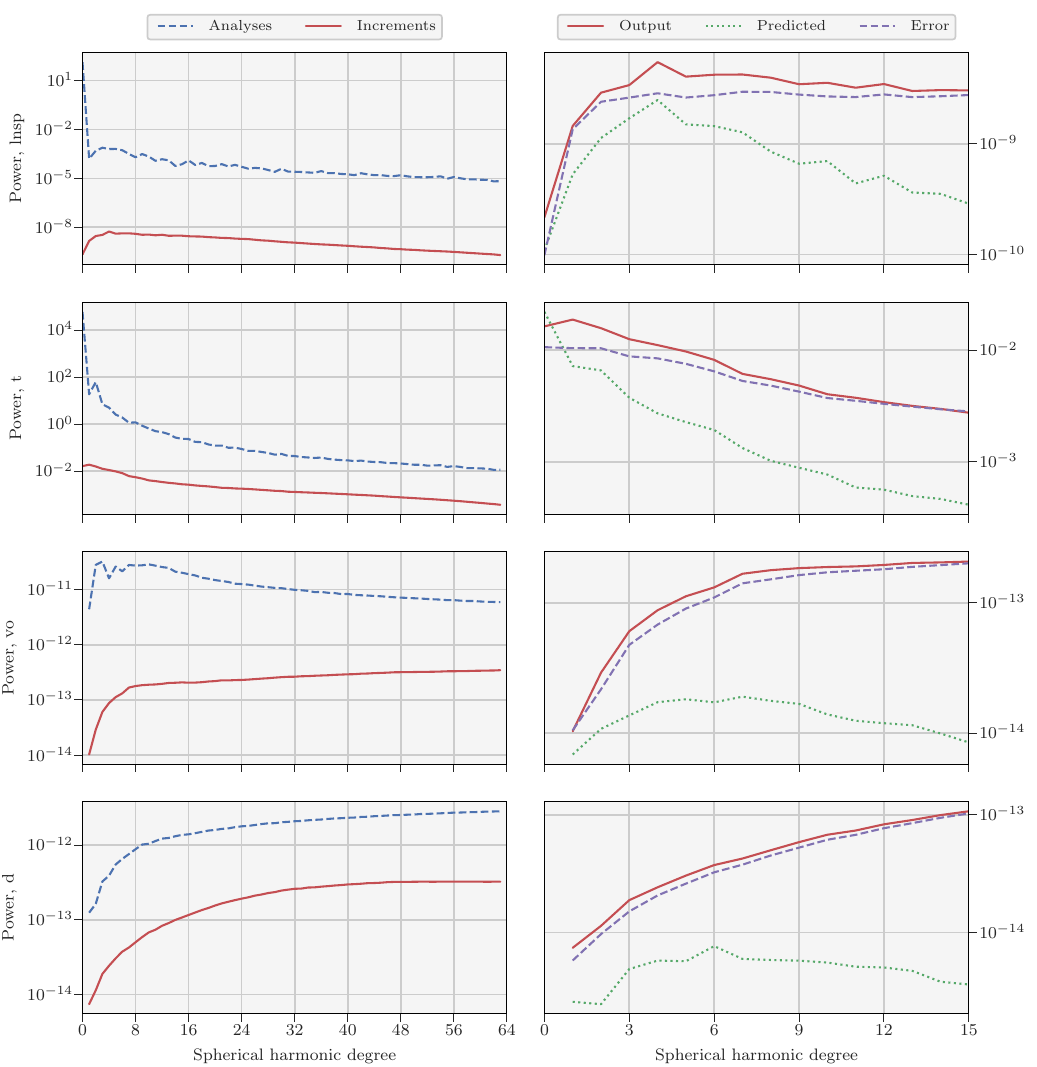}
    \caption{Averaged power spectra of the inputs (the analyses, in blue), the expected outputs (the increments, in red), the predictions (in green), and the prediction errors (difference between the expected and predicted increments, in purple) for each of the four variables: lnsp (first row), t (second row), vo (third row) and d (fourth row). For clarity, each row is split into two panels. The left panel shows the spectra of the inputs and expected outputs up to spectral degree $\num{63}$ (which is the resolution at which the data is available in the present work). The right panel shows the spactra of the expected outputs, the predictions, and the prediction error, up to spectral degree $\num{15}$ (which is the resolution at which the data is used in this experiment). Note that the red curve is exactly the same in both columns. Furthermore, for vo and d, the spectrum for spectral degree $\num{0}$ is not shown since it is supposed to be zero, to numerical precision.}
    \label{fig:offline-spectrum}
\end{figure}

Here, we compute one set of spectra for each variable, model level, and snapshot. The results are averaged over model levels and snapshots, in such a way that we end up with one set of spectra per variable. \add{Nevertheless, one should keep in mind that model levels are not evenly distributed in the vertical direction. These spectra are therefore more representative of the lowest atmosphere, which is much more represented in model levels, than the upper atmosphere.} The results are illustrated in \cref{fig:offline-spectrum}.

Overall, the analyses for lnsp and t are dominated by large scales (i.e. more power at low spectral degree) while the analyses for vo and d are characterised by smaller scales. A similar tendency can be observed for the analysis increments. By contrast, the predicted increments for all four variables are dominated by large scales, which is not a surprise since we expected large scale patterns to be more predictable than smaller scales patterns. Furthermore, the spectra of the predicted increments for all four variables are consistently much lower than the spectra of the actual increments, which is consistent with our previous finding that the neural network significantly underestimates the spatial variance of the increments (because the spatial variance is the sum of the power spectrum over all spectral degrees larger than $\num{1}$).

\begin{figure}[tbp]
    \centering
    \includegraphics[width=0.5\linewidth]{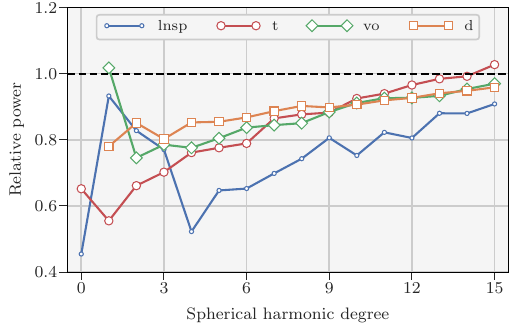}
    \caption{Relative averaged power spectra of the prediction errors for lnsp (in blue), t (in red), vo (in green), and d (in yellow).}
    \label{fig:offline-spectrum-rel}
\end{figure}

Consequently, for all four variables the spectra of the prediction errors are almost indistinguishable from the spectra of the actual increments. For this reason, we also illustrate in \cref{fig:offline-spectrum-rel} the relative spectra of the error, in other words for each variable the spectrum of the prediction errors divided by the spectrum of the true increments. Overall, the relative power of the error tends to increase with the spectral degree, which indicates that the neural networks estimations are more accurate at larger scales. There are however some exceptions for spectral degree $l\leq \num{3}$, which most probably corresponds to sampling noise. Indeed at a given spectral degree $l$, the power spectrum aggregates the contribution of \edit{$2\times(l+1)$}{$2\times l +1$} modes, which smoothes out the result.

\section{Towards higher resolution in offline training}
\label{app:offline-hr}

\add{[This section has moved to the appendix]} \edit{To close this section about offline training}{In this appendix}, we show the effect of increasing or decreasing the resolution of the training and validation data compared to the \edit{reference configuration which was used in the previous sections}{configuration used in \cref{sec:offline}}. Two setups are tested here: (i) a multivariate setup, with all four variables (lnsp, t, vo, and d) in the dataset, the same setup as in the reference configuration, and (ii) a univariate setup with only one variable (t in this case) in the dataset \add{for both the input and the output of the neural network}. For each setup, we use three resolutions:
\begin{itemize}
    \item T31 resolution, interpolated on the $\num{32}\times\num{63}$ Gauss--Legendre grid;
    \item T15 resolution, interpolated on the $\num{16}\times\num{31}$ Gauss--Legendre grid, the same resolution as in the reference configuration;
    \item T7 resolution, interpolated on the $\num{16}\times\num{31}$ Gauss--Legendre grid (i.e. the same grid as for the T15 resolution).
\end{itemize}
Note that in the T7 resolution, the data in grid-point space is widely over-sampled (the smallest Gauss--Legendre grid that can represent a field in the T7 resolution has $\num{8}\times\num{15}$ nodes). In all six cases, we train the exact same neural network as in the reference configuration, but only in prediction mode. In order to make a fair comparison between all resolutions, we choose to test the trained neural networks using the data in the original T63 resolution, interpolated on the $\num{64}\times\num{127}$ Gauss--Legendre grid.

\begin{table}[tbp]
    \centering
    \caption{Relative wMSE computed over the testing set (score $\mathcal{S}$) at T63 resolution.}
    \label{tab:offline-da-results-t63}
    \sisetup{round-mode=places, round-precision=3}
    \begin{tabular}{lllrrrrr}
    \headrow
    &&&\multicolumn{4}{c}{Multivariate} & Univariate \\
    \headrow
    \thead{Correction mode} & \thead{Resolution} & \thead{Grid} & $\mathcal{S}_{\mathsf{lnsp}}$ & $\mathcal{S}_{\mathsf{t}}$ & $\mathcal{S}_{\mathsf{vo}}$ & $\mathcal{S}_{\mathsf{d}}$ & $\mathcal{S}_{\mathsf{t}}$ \\
    Zero correction & --- & --- & \num{1} & \num{1} & \num{1} & \num{1} & \num{1} \\
    Prediction & T7 & $16\times31$  & \num{0.9993637800216675} & \num{0.981837809085846} & \num{1.0005649328231812} & \num{0.9996834993362427} & \num{0.8554102778434753} \\
    Prediction & T15 & $16\times31$  & \num{0.9279714226722717} & \num{0.8773393034934998} & \num{0.9934656023979187} & \num{0.9965654611587524} & \num{0.8421397805213928} \\
    Prediction & T31 & $32\times63$  & \num{0.8590449690818787} & \num{0.837268054485321} & \num{0.9844461679458618} & \num{0.9911220669746399} & \num{0.8433905839920044} \\
    \hline
    \end{tabular}
\end{table}

The global relative wMSE values are reported in \cref{tab:offline-da-results-t63}. The multivariate setup at T15 resolution corresponds to the reference configuration, but the scores are much higher here than in \cref{tab:offline-da-results} (where they were evaluated at T15 resolution), which is expected because we now have many more spectral degrees in the testing data. The loss of accuracy of the network \add{between evaluating at T15 (\cref{tab:offline-da-results}) and evaluating at T63 (\cref{tab:offline-da-results-t63}) }seems more important for vo and d than for lnsp and t, which is related to the fact that the increments for lnsp and t are dominated by larger scales than those for vo and d. In the end, as was concluded at T15 resolution, the increments for lnsp and t are more predictable than for vo and d, which are barely predictable. 

In the multivariate setup, a clear tendency emerges: the higher the resolution of the training data, the better the accuracy of the neural network for all four variables. In that case, the increased accuracy is not a direct consequence of the increase in the \edit{size of the training dataset}{number of training samples} (because in the T7 and T15 cases, we have used the exact same grid and hence the exact same \edit{training dataset size}{number of training samples}) but it is indeed a consequence of the increase in the resolution\add{, and hence in the information content,} of the dataset. By contrast, in the univariate setup there is little to no improvement when increasing the resolution of the training data. For all these reasons, we conclude that the cross-variables relationships depend on the resolution of the data. In the multivariate setup, when the neural network is trained at coarse resolution, it relies on cross-variables relationships which become inadequate when the network is tested at higher resolution. In our setup, this has a significant impact because variables such as vo and d have a strong signal at high resolution.

\begin{figure}[tbp]
    \centering
    \includegraphics[width=0.5\linewidth]{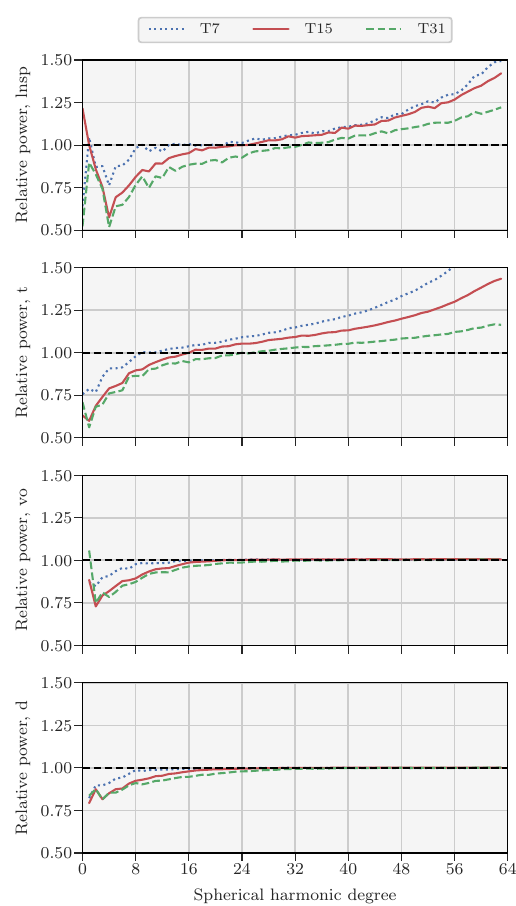}
    \caption{Relative averaged power spectra of the prediction errors for lnsp (top left panel), t (bottom left panel), vo (top right panel), and d (bottom right panel) when training at T7 (in blue), T15 (in red), and T63 (in green) resolution in the multivariate setup.}
    \label{fig:offline-spectrum-rel-t63}
\end{figure}

\begin{figure}[tbp]
    \centering
    \includegraphics[width=0.5\linewidth]{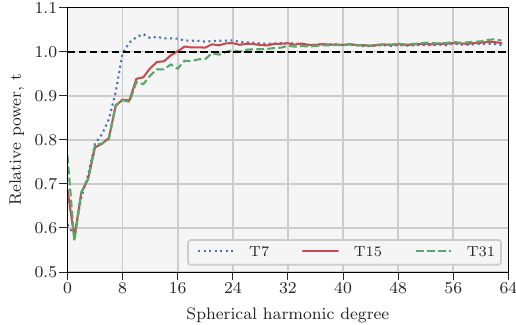}
    \caption{Relative averaged power spectra of the prediction errors for t when training at T7 (in blue), T15 (in red), and T63 (in green) resolution in the univariate setup.}
    \label{fig:offline-spectrum-rel-t63_t}
\end{figure}

To further illustrate the effect of resolution, we compute the relative averaged error power spectra (as defined in \cref{sapp:offline-spectral-analysis}). The results are shown in \cref{fig:offline-spectrum-rel-t63} for the multivariate setup and in \cref{fig:offline-spectrum-rel-t63_t} for the univariate setup. 

First, in the multivariate setup, when we look at the relative spectra of the neural network trained at T15 resolution (red lines in \cref{fig:offline-spectrum-rel-t63}), we observe similarities, but also differences compared to the spectra shown in \cref{fig:offline-spectrum-rel}. These differences arise because the test dataset in this case is at resolution T63, which includes spectral degrees higher than $\num{15}$. As the neural network operates as a nonlinear function in grid-point space, it is not expected to produce the exact same error spectrum as in \cref{fig:offline-spectrum-rel}. 

Second, in both mutlivariate and univariate cases, the relative power remains below one at spectral degrees lower than the training resolution. Conversely, the relative power tends to be above one or close to one at spectral degrees higher than the training resolution. This indicates that the neural network is able to correct errors only at a resolution lower than or equal to the one it was trained on.

Third, in the multivariate case, increasing the training resolution leads to decreased errors across nearly all spectral degrees. Conversely, in the univariate setup, increasing the training resolution reduces the errors only at high spectral degrees. This support our previous claim that, in the multivariate case, the decline in accuracy when training at lower resolution stems from the neural network depending on inadequate cross-variables relationships. 

Finally, note that at high spectral degrees the neural network often exhibits an increase in errors. This is particularly evident, in the multivariate setup, for lnsp and t. Interestingly, the increase in errors at high resolution for t is much less pronounced in the univariate setup compared to the multivariate setup, suggesting once again that it may be due to the neural network relying on inadequate cross-variables relationships. However, keep in mind that for lnsp and t the increase in errors is not as important as it may appear: at high resolution, the \edit{power}{spectral energy} of the increments is very low (as can be seen in \cref{fig:offline-spectrum}) which means that this increase has a limited effect on the predicted increments.

In the end, the multivariate setup has more potential than the univariate setup, because in the former case, the neural network can rely on cross-variables relationships to increase the accuracy of the predictions. However, the drawback of the multivariate setup is that, once the network is trained at a given resolution, using a different (e.g., higher) resolution later on may not be possible.

% \printendnotes

\bibliography{bibtex}

\graphicalabstract{fig_1}{In this article, we develop a neural network based model error correction for the operational Integrated Forecasting System (IFS). The neural network is pre-trained offline using a dataset of operational analyses and analysis increments and then online using a new variant of weak constraint 4D-Var. The results show that the network provides a reliable model error correction, which translated into reduced forecast errors in many conditions.}

\end{document}